\documentclass{article}

\usepackage{microtype}
\usepackage{graphicx}
\usepackage{subfig}
\usepackage{booktabs} % for professional tables
\usepackage{multirow}

\usepackage{hyperref}

\usepackage[accepted]{icml2024}

\usepackage{amsmath}
\usepackage{amssymb}
\usepackage{mathtools}
\usepackage{amsthm}
\usepackage{mathrsfs}

\allowdisplaybreaks[4]

\usepackage[capitalize,noabbrev]{cleveref}

\theoremstyle{plain}
\newtheorem{theorem}{Theorem}[section]
\newtheorem{proposition}[theorem]{Proposition}

\newtheorem{corollary}[theorem]{Corollary}
\theoremstyle{definition}
\newtheorem{definition}[theorem]{Definition}

\theoremstyle{remark}

\usepackage[textsize=tiny]{todonotes}

\icmltitlerunning{Invariant Risk Minimization Is A Total Variation Model}

\begin{document}

\def \bbE {\mathbb E}
\def \bbI {\mathbb I}
\def \bbR {\mathbb R}
\def \bbN {\mathbb N}
\def \bbV {\mathbb V}
\def \bbZ {\mathbb Z}

\def \mA {\mathcal{A}}
\def \mB {\mathcal{B}}
\def \mD {\mathcal{D}}
\def \mE {\mathcal{E}}
\def \mF {\mathcal{F}}
\def \mG {\mathcal{G}}
\def \mI {\mathcal{I}}
\def \mH {\mathcal{H}}
\def \mL {\mathcal{L}}
\def \mM {\mathcal{M}}
\def \mN {\mathcal{N}}
\def \mO {\mathcal{O}}
\def \mP {\mathcal{P}}
\def \mQ {\mathcal{Q}}
\def \mR {\mathcal{R}}
\def \mS {\mathcal{S}}
\def \mT {\mathcal{T}}
\def \mU {\mathcal{U}}
\def \mX {\mathcal{X}}
\def \mY {\mathcal{Y}}
\def \mZ {\mathcal{Z}}

\def \ff {\mathfrak{f}}

\def \sF {\mathscr{F}}
\def \sL {\mathscr{L}}
\def \sS {\mathscr{S}}

\def \ba {\bm{a}}
\def \bb {\bm{b}}
\def \bc {\bm{c}}
\def \bd {\bm{d}}
\def \bh {\bm{h}}
\def \bp {\bm{p}}
\def \bq {\bm{q}}
\def \bx {\bm{x}}
\def \by {\bm{y}}
\def \bz {\bm{z}}
\def \bw {\bm{w}}
\def \bu {\bm{u}}
\def \bv {\bm{v}}
\def \br {\bm{r}}
\def \bs {\bm{s}}
\def \bR {\bm{R}}
\def \bS {\bm{S}}
\def \bI {\bm{I}}
\def \bA {\bm{A}}
\def \bB {\bm{B}}
\def \bC {\bm{C}}
\def \bD {\bm{D}}
\def \bE {\bm{E}}
\def \bF {\bm{F}}
\def \bG {\bm{G}}
\def \bH {\bm{H}}
\def \bP {\bm{P}}
\def \bQ {\bm{Q}}
\def \bR {\bm{R}}
\def \bU {\bm{U}}
\def \bV {\bm{V}}
\def \bW {\bm{W}}
\def \bX {\bm{X}}
\def \bY {\bm{Y}}

\def \bone {\mathbf{1}}

\def \blambda {\bm{\lambda}}

\def \fp {\mathfrak{p}}
\def \frs {\mathfrak{s}}

\def \bbRd {\bbR^{d}}
\def \bbRn {\bbR^{n}}
\def \bbRm {\bbR^{m}}
\def \bbRdmY {\bbR^{d_{\mY}}}
\def \bbRdmH {\bbR^{d_{\mH}}}
\def \bbRE {\bbR^{E}}
\def \bbRN {\bbR^{N}}
\def \bbRM {\bbR^{M}}
\def \bbRT {\bbR^{T}}
\def \bbNM {\bbN_M}
\def \bbNk {\bbN_k}
\def \bbNn {\bbN_n}
\def \bbNm {\bbN_m}

\def \tS {\tilde{S}}
\def \tbQ {\tilde{\bQ}}
\def \tbq {\widetilde{\bq}}
\def \tbI {\widetilde{\bI}}

\def \st {\text{s.\ t.}}
\def \and {\text{and}}
\def \tr {\mathrm{tr}}
\def \prox {\mathrm{prox}}
\def \env {\mathrm{env}}
\def \supp {\mathrm{supp}}
\def \sign {\mathrm{sign}}
\def \trun {\mathrm{trun}}
\def \dist {\mathrm{dist}}
\def \div {\mathrm{div}}
\def \diag {\mathrm{diag}}
\def \Fix {\mathrm{Fix}}
\def \VaR {\mathrm{VaR}}
\def \CVaR {\mathrm{CVaR}}
\def \gra {\mathrm{gra}\hspace{2pt}}
\def \dom {\mathrm{dom}\hspace{2pt}}
\def \crit {\mathrm{crit}\hspace{2pt}}

\def \Argmax {\mathop{\rm Arg\,max}}
\def \Argmin {\mathop{\rm Arg\,min}}

\def \argmax {\mathop{\rm arg\,max}}
\def \argmin {\mathop{\rm arg\,min}}

\newcommand\leqs{\leqslant}
\newcommand\geqs{\geqslant}

\newcommand{\ud}{\,\mathrm{d}}
\def \dmu {\ud\mu}
\def \dnu {\ud\nu}
\def \drho {\ud\rho}

\twocolumn[
\icmltitle{Invariant Risk Minimization Is A Total Variation Model}




\begin{icmlauthorlist}
\icmlauthor{Zhao-Rong Lai}{yyy}
\icmlauthor{Weiwen Wang}{yyy}
\end{icmlauthorlist}

\icmlaffiliation{yyy}{Department of Mathematics, College of Information Science and Technology, Jinan University, Guangzhou, China}

\icmlcorrespondingauthor{Weiwen Wang}{wangww29@jnu.edu.cn}

\icmlkeywords{Invariant risk minimization, total variation model, coarea formula, out-of-distribution generalization}

\vskip 0.3in
]



\printAffiliationsAndNotice{}  % leave blank if no need to mention equal contribution

\begin{abstract}
Invariant risk minimization (IRM) is an arising approach to generalize invariant features to different environments in machine learning. While most related works focus on new IRM settings or new application scenarios, the mathematical essence of IRM remains to be properly explained. We verify that IRM is essentially a total variation based on $L^2$ norm (TV-$\ell_2$) of the learning risk with respect to the classifier variable. Moreover, we propose a novel IRM framework based on the TV-$\ell_1$ model. It not only expands the classes of functions that can be used as the learning risk and the feature extractor, but also has robust performance in denoising and invariant feature preservation based on the coarea formula. We also illustrate some requirements for IRM-TV-$\ell_1$ to achieve out-of-distribution generalization. Experimental results show that the proposed framework achieves competitive performance in several benchmark machine learning scenarios.
\end{abstract}

\section{Introduction}
\label{sec:intro}
Many machine learning tasks can be reduced to minimizing the learning risk, which is properly designed according to the task. Specifically, the learning risk is empirically minimized on the training set, but the learned model should be used on the unseen data. If there are significant differences between the features of the training and unseen data, the performance of the learning system may deteriorate \cite{imagenetgen,IRM,ZIN}. For example, if a convolutional neural network (CNN) is trained with pictures of cows in grass and camels in deserts, it may fail to classify easy samples of cows in deserts. This is because the CNN minimizes the training error by classifying grass and deserts as cows and camels, respectively. 
 
The key to solving this misclassification problem is to distinguish between the invariant features (animal shapes) and the spurious features (landscapes) which lead to distributional shift. Invariant Risk Minimization (IRM, \cite{IRM}) emerges as a learning paradigm that estimates nonlinear, invariant, and causal predictors from multiple training environments for such out-of-distribution (OOD) generalization issue. It introduces a gradient norm penalty that measures the optimality of the dummy classifier at each environment. Different variants of IRM have been proposed since then, such as Risk Extrapolation (REx, \citealt{REx}), SparseIRM \cite{SPIRM}, jointly-learning with auxiliary information (ZIN, \citealt{ZIN}), and invariant feature learning through independent variables (TIVA, \citealt{TIVA}). These works improve the robustness and extendability of IRM.

In this paper, we investigate the mathematical essence of IRM and find that it can be absorbed in a traditional and widely-used operator in mathematics: \textbf{total variation (TV)}. TV has long been used in various fields of mathematics and engineering, like optimal control, data transmission, sensing and denoising. For example, TV based on $\ell_1$ norm (TV-$\ell_1$, \cite{TVL1}) is exploited to develop noise removal algorithms in signal processing. TV-$\ell_1$ can also be used in image restoration and processing \cite{TVL1restore,LTV}. In other aspects, TV-$\ell_1$ is also a tractable approach to data-driven scale selection and function approximation \cite{tvl1appro}. 

Based on this motivation, we reveal some useful mathematical properties of TV-based IRM models that can be exploited in machine learning scenarios. \textbf{1.} We formulate and verify that the existing IRM framework is essentially a TV-$\ell_2$ model. \textbf{2.} We propose a novel IRM framework based on the TV-$\ell_1$ model (IRM-TV-$\ell_1$). It has two advantages: 1) The set of TV-$\ell_1$ integrable functions is larger than that of TV-$\ell_2$ integrable functions when the measure is finite (this is the case in IRM-TV-$\ell_2$), hence more classes of functions are allowed as the learning risk and the feature extractor in IRM-TV-$\ell_1$. 2) TV-$\ell_1$ has robust performance in denoising and preserving sharp features based on the coarea formula. This property helps to shape a blocky (piecewise-constant) learning risk that is more robust to the environment change. \textbf{3.} We investigate some requirements for IRM-TV-$\ell_1$ to achieve OOD generalization, such as a more flexible penalty parameter, extendability of the training environment set, and accuracy of the measure. These findings help to explain why IRM is effective theoretically.

\section{Preliminaries and Related Works}
\label{sec:prelim}
Note that there are different formulations and interpretations for both IRM and TV models. For the purpose of understanding the proposed approach, we adopt the most convenient formulations without loss of generality.

\subsection{Invariant Risk Minimization}
\label{sec:IRM}
Suppose we have a training data set of $n$ samples $\mD:=\{(x_i,y_i)\in \mX\times \mY\}_{i=1}^n$, where $\mX$ and $\mY$ denote the input space and the output space, respectively. These samples are collected from some training environments in the set $\mE_{tr}$ (e.g., grass or deserts). Note that we take a general scenario where environment partition is absent, thus the true environment label $e$ for a sample $(x_i,y_i)$ is unknown. The machine learning task is to learn a predictor $\mX\rightarrow \mY$ such that some predefined risk (or error) metric $R$ can be minimized on the training data $\mD$. In the IRM framework \cite{IRM}, this predictor is composed of two operators $w\circ \Phi$, where $\Phi: \mX\rightarrow \mH$ is able to extract invariant features and $w: \mH\rightarrow \mY$ is a classifier. 

Suppose we use the loss function $\mL: \mY\times\mY\times\mE_{tr}\rightarrow \bbR$ to compute the prediction error, where the environment variable becomes an argument of $\mL$. The empirical risk of the training samples in the environment $e$ is computed by
\begin{equation}
\label{eqn:emprisk}
R(w\circ \Phi,e):= \frac{1}{n}\sum_{i=1}^n \mL(w\circ \Phi(x_i),y_i,e),
\end{equation}
which is the mean loss between the predicted value $w\circ \Phi(x_i)$ and the true value $y_i$ in the environment $e$. The notations $x_i$ and $y_i$ can be omitted in $R$ for concise expressions, as they are constants. The actual parameters that we should learn are $w$, $\Phi$, and $e$.

The original IRM \cite{IRM} can be defined by
\begin{align}
\label{eqn:IRMorg}
\min_{w,\Phi}\ &\sum_{e\in \mE_{tr}} R(w\circ \Phi,e) \nonumber\\
\st \ & w \in \argmin_{\tilde{w}}  R(\tilde{w}\circ \Phi,e), \quad \forall e\in \mE_{tr}. \tag{IRM}
\end{align}
It aims to learn $w$ and $\Phi$ simultaneously by minimizing the total risk and forcing $w$ to be a uniform minimizer of the risk under each environment. However, since it is a challenging bi-level optimization problem, a practical surrogate can be used instead \cite{IRM}:
\begin{equation}
\label{eqn:IRMv1}
\min_{\Phi}\ \sum_{e\in \mE_{tr}} \left\{  R(1\circ \Phi,e) +\lambda  \|\nabla_w\arrowvert_{w=1} R(w\circ \Phi,e)\|_2^2 \right \}, \tag{IRMv1}
\end{equation}
where $\Phi$ becomes the entire invariant predictor and the classifier $w$ is fixed to be a scalar $1$. The penalty parameter $\lambda$ is non-negative, and the $\ell_2$ norm is denoted by $\|\cdot\|_2$. The minimization in the constraint of \eqref{eqn:IRMorg} is converted into a gradient norm penalty in the objective of \eqref{eqn:IRMv1}.

The REx \cite{REx} approach extrapolates the training risks to their affine combinations. The Variance-REx (V-REx) is a simple, stable and effective form of REx that takes the variance of training risks in different environments as the regularizer:
\begin{equation}
\label{eqn:VREx}
\min_{\Phi}\ \sum_{e\in \mE_{tr}}   R(1\circ \Phi,e) +\lambda  \bbV (\{R(1\circ \Phi,e)\}_{e\in \mE_{tr}}), \tag{V-REx}
\end{equation}
where $\bbV(\cdot)$ denotes the variance operator. The scalar $w=1$ is retained to be consistent with \eqref{eqn:IRMv1}. 

As for other extensions of IRM, ZIN \cite{ZIN} further exploits additional auxiliary information $\{z_i\in \mZ\}_{i=1}^n$ to simultaneously learn environment partition and invariant representation. It assumes that the training environment space $\mE_{tr}$ is a convex hull of $E$ linearly-independent fundamental environments (basis). Then each environment in $\mE_{tr}$ is isomorphic to an element in the $E$-dimensional simplex 
\begin{align}
\label{eqn:Esimplex}
\Delta^E:=\{v\in\bbRE: v\geqs 0_{(E)} \ \text{and} \ v\cdot 1_{(E)}=1  \},
\end{align}
where $0_{(E)}$ and $1_{(E)}$ denote $E$-dimensional vectors of $0$ and $1$, respectively. The inner product of two vectors is denoted by $\cdot$. ZIN aims to learn a mapping $\rho: \mZ\rightarrow \Delta^E$ that directly serves as the weights for different environments:
\begin{align}
\label{eqn:ZIN}
&R(w\circ \Phi,\rho):=  \frac{1}{n}\sum_{i=1}^n \tilde{\mL}(w\circ \Phi(x_i),y_i)\rho(z_i)\in \bbRE,\nonumber\\
&\min_{w,\Phi}\  \left\{  R(w{\circ} \Phi,\frac{1}{E}1_{(E)}){\cdot} 1_{(E)} {+}\lambda \max_{\rho} \|\nabla_w R(w{\circ} \Phi,\rho)\|_2^2 \right \}. \tag{ZIN}
\end{align}
Compared with \eqref{eqn:emprisk}, \eqref{eqn:ZIN} decomposes the environment argument into a convex combination of $E$ components. The treatment of the environment lies in learning $\rho$. $\frac{1}{E}1_{(E)}$ denotes the averaged environment, and $R(w\circ \Phi,\frac{1}{E}1_{(E)})\cdot 1_{(E)}$ measures the holistic empirical loss across all environments. The variation $\nabla_w R(w\circ \Phi,\rho)\in \bbR^{E\times d_{\mH}}$ becomes a gradient matrix, and the matrix $\ell_2$ norm is used as a penalty. The penalty is maximized w.r.t. $\rho$ to identify the worst environment that causes the largest absolute variation. Then the whole risk and penalty are minimized w.r.t. $(w,\Phi)$.

TIVA \cite{TIVA} directly uses the auxiliary information as input variables $\{\tilde{x}_i:=(x_i,z_i)\in\mX\times \mZ\}_{i=1}^n$ instead of sending it to the environment predictor $\rho$. It decomposes $\tilde{x}:= (\tilde{x}\odot u^+) + (\tilde{x}\odot (1-u^+))$ with a gate function $u^+\in [0,1]^{d_{\mX\times \mZ}}$ to identify environmental-related features. The risk metric of TIVA is defined by
\begin{align}
\label{eqn:TIVA}
&R(w{\circ} \Phi,\rho):=  \frac{1}{n}\sum_{i=1}^n \tilde{\mL}(w{\circ} \Phi(\tilde{x}{\odot} u^+),y_i)\rho(\tilde{x}{\odot} (1{-}u^+)). \tag{TIVA}
\end{align}
The optimization model is similar to \eqref{eqn:ZIN}.

\subsection{Total Variation}
\label{sec:TV}

Total variation (TV, \citealt{TVL1}) is a useful operator that characterizes the varying property of a function. Let $\Omega\subseteq \bbRd$ be an open set and $L^1(\Omega)$ be the space of functions defined on $\Omega$ with bounded Lebesgue integrals. Then for any function $f\in L^1(\Omega)$, its TV can be defined by \cite{TVL1restore}
\begin{align}
\label{eqn:TV}
\int_\Omega |\nabla f|:=\sup  \{  \int_\Omega & f(x)\div g(x)\ud x: \  g\in C_c^1(\Omega,\bbRd),\nonumber\\
& \|g\|_{L^\infty(\Omega)}\leqs 1    \}, \tag{TV}
\end{align}
where $g$ is a differentiable vector function with compact support contained in $\Omega$ and essential supremum no larger than $1$. $\div g$ denotes the divergence of $g$. In brief, the TV operator measures the local change of $f$ along all dimensions of $x$, then integrates all these local changes throughout the whole domain $\Omega$. Note that $f$ is not necessarily differentiable and the notation $|\nabla f|$ is symbolic. But if $f$ is differentiable, $|\nabla f|$ truly becomes the modulus (i.e., $\ell_2$ norm) of the gradient.

Only functions with bounded TV are meaningful in this realm, leading to the bounded variation (BV) space $\{f\in  L^1(\Omega): \int_\Omega |\nabla f|<\infty\}$. To simplify expressions, the term TV function actually means the BV function in this paper. In general, a TV model aims to minimize TV by default without explicitly stating it, because it makes little sense to maximize TV in practice. 

The TV operator is known for preserving sharp discontinuities while removing noise and other unwanted fine scale detail of $f$. This property is based on the coarea formula \cite{TVL1restore}
\begin{align}
\label{eqn:coarea}
\int_\Omega |\nabla f|:=\int_{-\infty}^{\infty}\int_{f^{-1}(\gamma)}\ud s \ud \gamma,
\end{align}
where $f^{-1}(\gamma):=\{x\in \Omega: f(x)=\gamma\}$ denotes the level set (preimage) of $f$ at $\gamma$. TV actually integrates along all contours of $f^{-1}(\gamma)$ for all $\gamma$ where the differential $\ud \gamma$ exists. Hence if $f$ is more blocky (piecewise-constant), it will have a smaller TV. This property is exploited in many signal recovery applications via the following TV-$\ell_1$ model \cite{TVL1}
\begin{align}
\label{eqn:TVL1sigrecover}
\inf_{f\in  L^2(\Omega)} \left\{ \int_\Omega |\nabla f|+ \lambda\int_\Omega (f-\tilde{f})^2\ud x \right\}, \tag{TV-$\ell_1$}
\end{align}
where $\tilde{f}\in  L^2(\Omega)$ is the ground-truth signal to be recovered and $\lambda$ is the approximating accuracy parameter. The objective of this model is to preserve sharp discontinuities in the approximation $f$ while leaving noise (especially Gaussian noise) and other unwanted fine scale detail in the residue $(f-\tilde{f})$.

If $f$ satisfies the variation condition $\int_\Omega |\nabla f|^2< \infty$, then the following TV-$\ell_2$ model \cite{TVL2} is feasible:
\begin{align}
\label{eqn:TVL2sigrecover}
\inf_{f\in  L^2(\Omega)} \left\{ \int_\Omega |\nabla f|^2+ \lambda\int_\Omega (f-\tilde{f})^2\ud x \right\}. \tag{TV-$\ell_2$}
\end{align}
In fact, TV-$\ell_2$ appears earlier in history than TV-$\ell_1$. However, the square-integral $\int_\Omega |\nabla f|^2$ does not have the coarea formula \eqref{eqn:coarea}. Therefore, TV-$\ell_2$ generally does not produce a blocky recovered signal $f$.

\section{Methodology}
\label{sec:IRMintoTVM}
Motivated by the properties of TV, we aim to develop a risk metric that is blocky (piecewise-constant) w.r.t. the environment, so that it can be better generalized to different environments. First, we establish the theoretical framework for IRM to be a TV-$\ell_2$ model.

\subsection{Conditions for IRM to be TV-$\ell_2$}
\label{sec:IRM-TVL2}
A necessary but not sufficient condition for the classifier $w$ to achieve the minimums in the constraints of \eqref{eqn:IRMorg} is
\begin{align}
\label{eqn:IRMorgness}
\nabla_w R(w\circ \Phi,e)=0,  \quad \forall e\in \mE_{tr}.
\end{align}
This leads to the surrogate model \eqref{eqn:IRMv1} which penalizes the gradient norms over all the environments. However, the risk function $R(w\circ \Phi,e)$ usually has a complex nonconvex geometric structure w.r.t. $w$ in deep learning. Thus \eqref{eqn:IRMorgness} may achieve a maximum instead of a minimum, and it is intractable to turn \eqref{eqn:IRMv1} back to \eqref{eqn:IRMorg}.

From a different perspective, we interpret \eqref{eqn:IRMv1} as a TV-$\ell_2$ model in two ways: 1. Consider the environment features as noise and try to remove it. 2. Recover the useful signal (e.g., animal shapes) from the background environment.

To begin with, we turn $w$ back to an argument in the minimization of \eqref{eqn:IRMv1}. This is reasonable because the learning of $w$ and $\Phi$ is a dynamic process. 
\begin{align}
\label{eqn:IRMv1w}
\min_{w,\Phi}\ &\sum_{e\in \mE_{tr}} \left\{  R(w\circ \Phi,e) +\lambda  \|\nabla_w R(w\circ \Phi,e)\|_2^2 \right \}. 
\end{align}
We also need the following successive conditions.

\textbf{Condition 1.} There exists a measure $\mu$ for $e$ on $(\mE_{tr},\sF_{tr})$ where $\sF_{tr}$ denotes the $\sigma$-algebra on $\mE_{tr}$ that includes all the environment combinations in the training set. 

\textbf{Condition 2.} Under this measure $\mu$, the feature extractor $\Phi$ is uncorrelated to the environment variable $e$. Moreover, the correlation of the risk metric $R$ to $e$ only lies on the classifier $w$, i.e., $R(w\circ \Phi,e)\dmu\equiv R(w(e)\circ \Phi)\dmu$.

\textbf{Condition 3.} $w$ is a measurable function of $e$ in the sense that $w$ is parameterized (either fully or partly) by $e$. In other words, $w(\cdot)\circ \Phi: \mE_{tr}\times \mH  \rightarrow \mY$ is measurable.

\textbf{Condition 4.} $R(w(e)\circ \Phi)\in L^1(\mE_{tr},\sF_{tr},\mu)$ and $|\nabla_w R(w(e)\circ \Phi)|\in  L^2(\mE_{tr},\sF_{tr},\mu)$. $L^1(\mE_{tr},\sF_{tr},\mu)$ and $L^2(\mE_{tr},\sF_{tr},\mu)$ denote the function spaces with integrable and square-integrable functions under measure $\mu$ defined on $(\mE_{tr},\sF_{tr})$, respectively. 

Condition 1 implies that we are able to measure the magnitude of the environments we encounter in the training set. For instance, \eqref{eqn:IRMv1w} actually uses a counting measure on $\mE_{tr}$ such that for any $e\in\mE_{tr}$, $\dmu(\{e\})=1$. Condition 2 is weaker than $R(w\circ \Phi,e)\equiv R(w(e)\circ \Phi)$ and indicates that the environment influences only $w$ but not $\Phi$ under an infinitesimal scale of environment change $\dmu$. Then $w$ passes the environment representation to $R$. This is reasonable because $\Phi$ is the invariant feature extractor that we intend to learn. Condition 3 enables us to represent and measure $w$ by $e$. Finally, Condition 4 is fundamental to ensure boundedness when we compute the total risk and its TV.

\begin{theorem}
\label{thm:IRMTVL2}
A risk metric $R(w\circ \Phi,e)$ that satisfies Conditions 1$\sim$4 has the following well-defined finite integral and TV-$\ell_2$ form w.r.t. $w$:
\begin{align}
\label{eqn:IRMintTVself}
&\int_{\Omega} R(w\circ \Phi) \dnu \quad \and \quad\int_{\Omega}   |\nabla_w R(w\circ \Phi)|^2 \dnu, 
\end{align}
where $\Omega=w(\mE_{tr})$ is the image of $\mE_{tr}$ under mapping $w$, and $\nu$ is the measure for $w$ induced by $\mu$. If $\mu$ is a probability measure, then the above integrals become mathematical expectations
\begin{align}
\label{eqn:IRMexpTVself}
&\bbE_{w} [R(w\circ \Phi)]  \quad \and \quad \bbE_{w}   [|\nabla_w R(w\circ \Phi)|^2].
\end{align}
The following IRM-TV-$\ell_2$ model is also well-defined:
\begin{align}
\label{eqn:IRMv1wTVL2}
\min_{\Phi}\ &\bbE_{w} [R(w\circ \Phi) +\lambda  |\nabla_w R(w\circ \Phi)|^2 ]. \tag{IRM-TV-$\ell_2$}
\end{align}
\end{theorem}
The proof is presented in Appendix \ref{proof:thmIRMTVL2}. Note that only $\Phi$ is left in the minimizing argument because $w$ has been taken expectation in the objective function. This reveals the mathematical essence of why $w$ can be set as a dummy scalar classifier $1$ in \eqref{eqn:IRMv1}. 

It follows from Theorem \ref{thm:IRMTVL2} and multiplying \eqref{eqn:IRMv1w} by $1/|\mE_{tr}|$ that the following corollary can be obtained.
\begin{corollary}
\label{cor:IRMv1w}
Let $\mE_{tr}$ be a finite discrete environment training set with a discrete uniform probability measure $\mu$.  Then given Conditions 1$\sim$4, \eqref{eqn:IRMv1w} can be generalized to \eqref{eqn:IRMv1wTVL2}.
\end{corollary}
By slightly changing Condition 4, \eqref{eqn:VREx} can also be generalized like Theorem \ref{thm:IRMTVL2}.

\textbf{Condition 4a.} $R(\Phi,e)\in L^2(\mE_{tr},\sF_{tr},\mu)$ with $\mu$ being a probability measure.

\begin{proposition}
\label{cor:VREx}
Given Conditions 1 and 4a, \eqref{eqn:VREx} can be generalized to the following V-REx-$\ell_2$ model:
\begin{align}
\label{eqn:VRExTVL2}
\min_{\Phi}\ &\bbE_{e} [R(\Phi,e)] {+}\lambda  \bbE_{e} [(R(\Phi,e){-}\bbE_{e} [R(\Phi,e)])^2]. \tag{V-REx-$\ell_2$}
\end{align}
\end{proposition}
The proof is presented in Appendix \ref{proof:VREx}. In \eqref{eqn:VRExTVL2}, $\mE_{tr}$ is not necessarily a finite discrete set and $\mu$ is not necessarily a uniform probability measure. Besides, \eqref{eqn:VRExTVL2} does not require a classifier $w$ to convey environment information (Conditions 2 and 3). This proposition reveals the mathematical essence of why V-REx can extrapolate risk to a larger region (see Appendix \ref{proof:VREx}). \eqref{eqn:VRExTVL2} can be considered as a variant of TV-$\ell_2$ model, as variance is a similar tool to TV that assesses the deviation of a function.

To incorporate ZIN and TIVA into our TV-$\ell_2$ framework, we present the following theorem.
\begin{theorem}
\label{prop:ZIN}
The environment learner $\rho$ is actually a probability measure satisfying Condition 1. Given Conditions 2$\sim$4, ZIN and TIVA can be generalized to the following Minimax-TV-$\ell_2$ model
\begin{align}
\label{eqn:MinimaxTVL2}
\min_{\Phi}\ &\{ \bbE_{w \leftarrow \frac{1_{(E)}}{E}} [ R(w{\circ} \Phi)] {+}\lambda  \max_{\rho}\bbE_{w\leftarrow \rho} [|\nabla_w R(w{\circ} \Phi)|^2 ]  \}, \tag{Minimax-TV-$\ell_2$}
\end{align}
where $\bbE_{w\leftarrow \rho}$ denotes the mathematical expectation w.r.t. $w$ whose measure $\nu$ is induced by $\rho$, and $\frac{1_{(E)}}{E}$ denotes the uniform probability measure for $e$.  
\end{theorem}
The proof is presented in Appendix \ref{proof:ZINL2}. This theorem indicates that this minimax scheme actually selects a probability measure $\rho$ that can maximize the risk variation (the worst case), then minimizes this worst-case total objective by $\Phi$.

\subsection{IRM-TV-$\ell_1$}
\label{sec:IRM-TVL1}
In last section, we verify that some typical IRM models and related extensions are essentially TV-$\ell_2$ models. However, IRM-TV-$\ell_2$ does not have the coarea formula \cite{TVL1restore} that provides a geometric nature of sharp discontinuity preservation, despite having the denoising property. As a result, the processed learning risk $R$ may still be environment-sensitive.

To address this issue, we establish a novel IRM-TV-$\ell_1$ framework that enhances the robustness of learning risk to environment changes based on the coarea formula. Since $L^2(\Omega,\sF_{\Omega},\nu)\subsetneqq L^1(\Omega,\sF_{\Omega},\nu)$ when $\nu(\Omega)<\infty$ (this is usually the case especially when $\nu$ is a probability measure, including \ref{eqn:IRMv1wTVL2} and \ref{eqn:MinimaxTVL2}), IRM-TV-$\ell_1$ allows broader classes of functions to be the learning risk $R$ and the feature extractor $\Phi$, which enables us to address more complicated tasks and explore more feature extractors, respectively. To achieve this, we can relax Condition 4 as the following Condition 4b. It is a necessary but not sufficient condition for Condition 4 when $\mu(E_{tr})<\infty$.

\textbf{Condition 4b.} $R(w(e)\circ \Phi), |\nabla_w R(w(e)\circ \Phi)|\in L^1(\mE_{tr},\sF_{tr},\mu)$.

\begin{theorem}
\label{thm:IRMTVL1}
$\Omega$, $\nu$, $\bbE_{w}$ and $\rho$ are the same as those in Theorems \ref{thm:IRMTVL2} and \ref{prop:ZIN}. A risk metric $R(w\circ \Phi,e)$ satisfying Conditions 1$\sim$3 and 4b has the following well-defined integral, TV-$\ell_1$ form and mathematical expectations w.r.t. $w$:
\begin{align}
\label{eqn:IRMintTVL1self}
&\int_{\Omega} R(w\circ \Phi) \dnu \quad \and \quad\int_{\Omega}   |\nabla_w R(w\circ \Phi)| \dnu, \\
\label{eqn:IRMexpTV1self}
&\bbE_{w} [R(w\circ \Phi)]  \quad \and \quad \bbE_{w}   [|\nabla_w R(w\circ \Phi)|].
\end{align}
The following IRM-TV-$\ell_1$ and Minimax-TV-$\ell_1$ models are also well-defined:
\begin{align}
\label{eqn:IRMv1wTVL1}
\min_{\Phi}\ &\left\{ \bbE_{w} [R(w\circ \Phi)] +\lambda  (\bbE_{w} [|\nabla_w R(w\circ \Phi)| ])^2 \right\}, \tag{IRM-TV-$\ell_1$}\\
\label{eqn:MinimaxTVL1}
\min_{\Phi}\ &\{ \bbE_{w \leftarrow \frac{1_{(E)}}{E}} [ R(w{\circ} \Phi)] {+}\lambda  \max_{\rho}(\bbE_{w\leftarrow \rho} [|\nabla_w R(w{\circ} \Phi)| ])^2  \}. \tag{Minimax-TV-$\ell_1$}
\end{align}
\end{theorem}
The proof is presented in Appendix \ref{proof:IRMTVL1}. Note that $(\bbE_{w} [|\nabla_w R(w\circ \Phi)| ])^2 =\|\nabla_w R(w\circ \Phi)\|_1^2$ if $\|\cdot\|_1$ is the $L^1$ norm for $L^1(\Omega,\sF_{\Omega},\nu)$, while $\bbE_{w} [  |\nabla_w R(w\circ \Phi)|^2 ]=\|\nabla_w R(w\circ \Phi)\|_2^2$ if $\|\cdot\|_2$ is the $L^2$ norm for $L^2(\Omega,\sF_{\Omega},\nu)$. Hence we retain a squared TV-$\ell_1$ term in \eqref{eqn:IRMv1wTVL1} to be consistent with the squared TV-$\ell_2$ term in \eqref{eqn:IRMv1wTVL2}, so as \eqref{eqn:MinimaxTVL1}. 

The term $|\nabla_w R(w\circ \Phi)|$ in \eqref{eqn:IRMv1wTVL1} and \eqref{eqn:MinimaxTVL1} is non-differentiable w.r.t. $\Phi$. Therefore, conventional backpropagation or other gradient-type methods cannot be used in this model. We develop a closed-form subgradient computation to solve \eqref{eqn:IRMv1wTVL1} and \eqref{eqn:MinimaxTVL1}, shown in Appendix \ref{sec:IRMTVL1solve}. It operates similarly to a gradient computation and will not increase computational complexity.

The following proposition ensures that IRM-TV-$\ell_1$ has the coarea formula, which produces a learning risk that is blocky (piecewise-constant) w.r.t. the environment. It explains the mathematical essence of why IRM-TV-$\ell_1$ is able to learn invariant features. 
\begin{proposition}
\label{prop:IRMTVL1coarea}
Assume the induced measure $\nu$ for $w$ to be the Lebesgue measure on $\bbRd$. Given Conditions 1$\sim$3 and 4b, the following coarea formula holds:
\begin{align}
\label{eqn:IRMTVL1coarea}
\int_\Omega |\nabla_w R(w\circ \Phi)| \dnu=\int_{-\infty}^{\infty}\int_{\{w\in \Omega: R(w\circ \Phi)=\gamma\}} \ud s \ud \gamma,
\end{align}
where $s$ is the $(d-1)$-dimensional Hausdorff measure (i.e., area in $(d-1)$ dimensions). 
\end{proposition}
The proof is given in Appendix \ref{proof:IRMTVL1coarea}.

\subsection{Out-of-distribution Generalization}
\label{sec:OOD}
OOD generalization \cite{OOD} refers to the ability of a trained model to be generalized to an unseen domain. It can be formally defined as follows \cite{IRM,groupDRO,DRO,REx,ZIN}:
\begin{align}
\label{eqn:OOD}
 &\min_{\Phi}\max_{e\in \mE_{all}} R(\Phi,e), \tag{OOD}
\end{align}
where $\mE_{all}$ denotes the global environment set that contains all the possible environments that can occur in the test, and $\Phi$ denotes the model parameters for $R$. Two facts are implied in this framework: 1. $\max_{e\in \mE_{all}} R(\Phi,e)<\infty$ for some $\Phi$, or else it cannot be minimized by any $\Phi$. 2. Both $\argmax_{e\in \mE_{all}} R(\Phi,e)$, $\forall \Phi$ and $\argmin_{\Phi}\max_{e\in \mE_{all}} R(\Phi,e)$ exist, or else the worst-case environment or the optimal model parameters cannot be identified. To absorb the environment variable $e$ into the classifier $w$, Condition 2 should be strengthened:

\textbf{Condition 2a.} The representation of the risk metric $R$ by $e$ only lies on the classifier $w$, i.e., $R(\Phi,e)\equiv R(w(e)\circ \Phi)$.

We assume that the following Theorems \ref{thm:OODwfrome}, \ref{thm:OODwequiv} and \ref{thm:OODwequivmm} satisfy Conditions 1,2a,3 and 4b on the probability space $(\mE_{all},\sF_{all},\mu_{all})$. 
\begin{theorem}
\label{thm:OODwfrome}
\eqref{eqn:OOD} is equivalent to
\begin{align}
\label{eqn:OODwfrome}
 &\min_{\Phi}\max_{w\in \Omega_{all}} R(w\circ \Phi), \tag{OOD-$w$}
\end{align}
where $\Omega_{all}=w(\mE_{all})$.
\end{theorem}
The proof is given in Appendix \ref{proof:OODwfrome}. In fact, IRM, ZIN and TIVA implicitly exploit this theorem to set up their models.

On the other hand, \eqref{eqn:IRMv1wTVL1} under the global environment set becomes
\begin{align}
\label{eqn:IRMv1wTVL1glob}
\min_{\Phi}\ &\left\{ \bbE_{w}^{all} [R(w\circ \Phi)] +\lambda  (\bbE_{w}^{all} [|\nabla_w R(w\circ \Phi)| ])^2 \right\}, \tag{IRM-TV-$\ell_1$-global}
\end{align}
where $\bbE_{w}^{all}$ denotes the expectation w.r.t. $w$ on the induced probability space $(\Omega_{all},\sF_{\Omega_{all}},\nu_{all})$. Our first step to establish OOD generalization is to investigate the conditions under which \eqref{eqn:IRMv1wTVL1glob} is really equivalent to \eqref{eqn:OODwfrome}.

\begin{theorem}[IRM-TV-$\ell_1$-global Achieving OOD Generalization]
\label{thm:OODwequiv}

1) The penalty parameter $\lambda$ should be allowed to vary with $\Phi$. Otherwise, \eqref{eqn:IRMv1wTVL1glob} cannot achieve \eqref{eqn:OODwfrome}.

2) For each $\Phi$, if $\bbE_{w}^{all} [|\nabla_w R(w{\circ} \Phi)| ]> 0$, then there exists some $\lambda_\Phi\geqs 0$ depending only on $\Phi$ such that 
\begin{align}
\label{eqn:OODwequal}
&\max_{w\in \Omega_{all}} R(w{\circ} \Phi)\nonumber\\
&{=} \bbE_{w}^{all} [R(w{\circ} \Phi)] {+}\lambda_\Phi  (\bbE_{w}^{all} [|\nabla_w R(w{\circ} \Phi)| ])^2.
\end{align}
If $\bbE_{w}^{all} [|\nabla_w R(w{\circ} \Phi)| ]= 0$, \eqref{eqn:OODwequal} still holds when $R(w{\circ} \Phi)$ is Lipschitz continuous w.r.t. $w$.

3) An optimal point $\Phi^\bullet$ of the following model
\begin{align}
\label{eqn:IRMv1wTVL1globphi}
\min_{\Phi}\ &\left\{ \bbE_{w}^{all} [R(w\circ \Phi)] +\lambda_\Phi  (\bbE_{w}^{all} [|\nabla_w R(w\circ \Phi)| ])^2 \right\} \tag{IRM-TV-$\ell_1$-global-$\Phi$}
\end{align}
is also an optimal point of \eqref{eqn:OODwfrome}, and vice versa.
\end{theorem}
The proof is given in Appendix \ref{proof:OODwequiv}. Item 1) allows $\lambda$ to vary with $\Phi$. Without this variation, even a simple function fitting task cannot be generalized, as shown in Appendix \ref{proof:OODwequiv}. Item 2) ensures the existence of $\lambda_\Phi$ that fills the gap between the maximum and the expectation of $R(w\circ \Phi)$. Item 3) establishes the equivalence between the optimal points of \eqref{eqn:IRMv1wTVL1glob} and \eqref{eqn:OODwfrome}.

\begin{theorem}[Minimax-TV-$\ell_1$-global Achieving OOD Generalization]
\label{thm:OODwequivmm}
1) The penalty parameter $\lambda$ should be allowed to vary with $\Phi$. Otherwise, the following model cannot achieve \eqref{eqn:OODwfrome}: 
\begin{align}
\label{eqn:MinimaxTVL1glob}
\min_{\Phi}\ &\{ \bbE_{w \leftarrow \frac{1_{(E)}}{E}}^{all} [ R(w{\circ} \Phi)] {+}\lambda  \max_{\rho}(\bbE_{w\leftarrow \rho}^{all} [|\nabla_w R(w{\circ} \Phi)| ])^2  \}. \tag{Minimax-TV-$\ell_1$-global}
\end{align}

2) For each $\Phi$, if $\bbE_{w\leftarrow \rho}^{all} [|\nabla_w R(w{\circ} \Phi)| ]> 0$ for some $\rho$, then there exists some $\lambda_\Phi\geqs 0$ depending only on $\Phi$ such that 
\begin{align}
\label{eqn:OODwequalmm}
&\max_{w\in \Omega_{all}} R(w{\circ} \Phi)\nonumber\\
&{=}\bbE_{w \leftarrow \frac{1_{(E)}}{E}}^{all} [ R(w{\circ} \Phi)] {+}\lambda_\Phi  \max_{\rho}(\bbE_{w\leftarrow \rho}^{all} [|\nabla_w R(w{\circ} \Phi)| ])^2.
\end{align}
If $\bbE_{w\leftarrow \rho}^{all} [|\nabla_w R(w{\circ} \Phi)| ]= 0$ for all $\rho$, \eqref{eqn:OODwequal} still holds when $R(w{\circ} \Phi)$ is Lipschitz continuous w.r.t. $w$.

3) An optimal point $\Phi^\bullet$ of the following model
\begin{align}
\label{eqn:MinimaxTVL1globphi}
\min_{\Phi}&\{\bbE_{w \leftarrow \frac{1_{(E)}}{E}}^{all} [ R(w{\circ} \Phi)] {+}\lambda_\Phi  \max_{\rho}(\bbE_{w\leftarrow \rho}^{all} [|\nabla_w R(w{\circ} \Phi)| ])^2 \} \tag{Minimax-TV-$\ell_1$-global-$\Phi$}
\end{align}
is also an optimal point of \eqref{eqn:OODwfrome}, and vice versa.
\end{theorem}
The proof is given in Appendix \ref{proof:OODwequivmm}.

Theorems \ref{thm:OODwfrome}, \ref{thm:OODwequiv}, and \ref{thm:OODwequivmm} reveal the conditions under which TV-$\ell_1$ models can achieve OOD generalization under the global environment set. We then investigate whether the training environment set can be generalized to the global environment set. In most cases, we encounter some fundamental environments in the training set that form a basis for environment representation, either explicitly (e.g., IRM, REx) or implicitly (e.g., ZIN, TIVA).
\begin{definition}[Basis from Training Environments]
\label{dfn:basisgeneral}
\begin{align}
\label{eqn:basisgeneral}
\mB_{tr}:=\{e_i\}_{i\in E},
\end{align}
where $E$ denotes the index set for this basis. $E$ can be finite, countable or uncountable.
\end{definition}
Normally, $\mB_{tr}\subseteq\mE_{all}$, and usually $\mB_{tr}\subsetneqq\mE_{all}$ because there may be unseen environments outside the training set. This is why OOD generalization methods try to expand $\mB_{tr}$ from different approaches. An intuitive and popular approach is to consider the linear space spanned by $\mB_{tr}$:
\begin{align}
\label{eqn:basisspan}
\mE_{tr}^{\sL}:=\sL(\mB_{tr}).
\end{align}
REx \cite{REx} explicitly takes this form for the training environment set (see Appendix \ref{proof:VREx}). Although ZIN and TIVA aim to expand $\mB_{tr}$ to its convex hull, they actually take $\mE_{tr}^{\sL}$ as well and adopt a probability measure $\rho$ on it according to Theorem \ref{prop:ZIN}.

\begin{figure}[h]
\centering
\includegraphics[width=0.9\columnwidth]{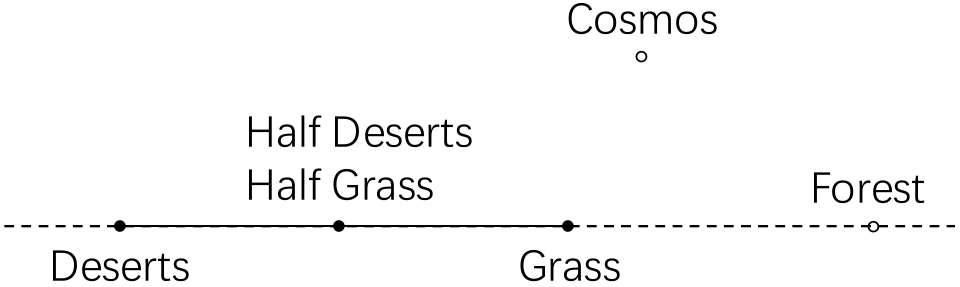}
\caption{A simple example for the linear environment space $\mE_{tr}^{\sL}$ (the dashed line). }
\label{fig:linenvsample}
\vspace{-10pt}
\end{figure}

The following theorem reveals the minimum requirements for global environment set generalization.
\begin{theorem}
\label{thm:OODminimum}
Under the same conditions as in Theorem \ref{thm:IRMTVL1}, $\mE_{all}\in \sF_{tr}$ is the minimum requirement for \eqref{eqn:IRMv1wTVL1} and \eqref{eqn:MinimaxTVL1} to be generalized to \eqref{eqn:IRMv1wTVL1globphi} and \eqref{eqn:MinimaxTVL1globphi}, respectively.
\end{theorem}
The proof is given in Appendix \ref{proof:OODminimum}. This theorem reveals two important facts:

\textbf{1. $\mE_{tr}$ should be abundant and diverse enough.} In particular, $\mE_{tr}\supseteq\mE_{all}$ is necessary. It may seem impossible at first thought, but one can try to expand $\mB_{tr}$ as demonstrated above, like $\mE_{tr}^{\sL}$. After all, a model can hardly foresee the environment of the cosmos in $\mE_{all}$ with only grass and deserts in $\mB_{tr}$. But it may be able to learn the environment with half deserts and half grass, or even the environment of a forest (see Figure \ref{fig:linenvsample}). The maximum $\sigma$-algebra of $\mE_{tr}$ is its power set $2^{\mE_{tr}}$. If $\mE_{all}\notin 2^{\mE_{tr}}$, then distortion is inevitable in  \eqref{eqn:IRMv1wTVL1globphi} and \eqref{eqn:MinimaxTVL1globphi}. Recently, the spurious feature diversification strategy \cite{spuriousfeat} has been proposed to expand $\mE_{tr}$, which is consistent with the above analysis.

\textbf{2. The measure $\mu$ for $e$ on $(\mE_{tr},\sF_{tr})$ should be accurate enough.} Even if $\mE_{tr}\supseteq\mE_{all}$ cannot guarantee a measurable $\mE_{all}$. For example, let $\mE_{tr}\supsetneqq\mE_{all}\supsetneqq\emptyset$ and the measure $\mu$ be the coarsest one such that $\sF_{tr}=\{\emptyset,\mE_{tr}\}$, then $\mu(\mE_{all})$ does not exist. For another example, let $\mE_{tr}=\bbRE$, then the most accurate measure that can be realized by modern computing technologies is the Lebesgue measure $\mu_{L}$ defined on the Lebesgue $\sigma$-algebra $\sF_{L}$. However, $\sF_{L}\subsetneqq2^{\bbRE}$ and there are non-measurable sets such as the Vitali set. Hence if $\mE_{all}\notin \sF_{L}$, distortion is also inevitable in \eqref{eqn:IRMv1wTVL1globphi} and \eqref{eqn:MinimaxTVL1globphi}. It makes little sense to only expand $\mE_{tr}$ without equipping it with an accurate $\mu$.

Theorem \ref{thm:OODminimum} actually deploys $\sF_{tr}$ as the $\sigma$-algebra for $\mE_{all}$. However, $\mE_{all}$ may have its own information structure, represented by its own $\sigma$-algebra $\sF_{all}$ and measure $\mu_{all}$. Then it will be more complicated to explore \eqref{eqn:OODwfrome}, leading to the following sufficient requirements.
\begin{corollary}
\label{cor:OODsufficient}
Under the same conditions as in Theorem \ref{thm:IRMTVL1}, $\sF_{tr}\supseteq\sF_{all}$ is the sufficient requirement for \eqref{eqn:IRMv1wTVL1} and \eqref{eqn:MinimaxTVL1} to be generalized to \eqref{eqn:IRMv1wTVL1globphi} and \eqref{eqn:MinimaxTVL1globphi}, respectively.
\end{corollary}
To summarize this subsection, there are three ways to improve OOD generalization: 1. Allow $\lambda_\Phi$ to vary with $\Phi$. 2. Expand the training environment basis $\mB_{tr}$ to a larger training environment set $\mE_{tr}$ by all means. 3. Equip $\mE_{tr}$ with a more accurate measure $\mu$.

\section{Experiments}

We assess the proposed IRM-TV-$\ell_{1}$ framework by comparisons with existing OOD methods in simulation and real-world experiments, following \cite{ZIN, TIVA}. Seven state-of-the-art methods: IRM \cite{IRM}, groupDRO \cite{groupDRO}, EIIL \cite{EIIL}, LfF \cite{LfF}, HRM \cite{HRM}, ZIN \cite{ZIN}, and TIVA \cite{TIVA}, as well as the baseline Empirical risk minimization (ERM) are taken into comparisons. IRM, groupDRO, and \ref{eqn:IRMv1wTVL1} require ground-truth environment partitions, while \ref{eqn:MinimaxTVL1} is the counterpart of \ref{eqn:IRMv1wTVL1} for  non-environment-partition situations. Implementation details are presented in Appendix \ref{implementation}. Each experiment is repeated for 10 times to compute the averaged result and the standard deviation (STD) for each method.

\subsection{Simulation Study}
Each synthetic data set is characterized by temporal heterogeneity with distributional shift w.r.t time and has been used for OOD generalization evaluation in \cite{ZIN, TIVA}. For time $t \in [0, 1]$, the interested binary outcome is denoted by $Y(t)$, whose cause and effect are the invariant features $X_{v}(t)\in \mathbb{R}$ and the spurious features $X_{s}(t) \in \mathbb{R}$, respectively (see Appendix \ref{imp:gensyn} for more details). The correlation between $X_{v}(t)$ and $Y(t)$ is stable and controlled by a parameter $p_{v}$, while the correlation between $X_{s}(t)$ and $Y(t)$ is changing over $t$ and controlled by a parameter $p_{s}(t)$. In training sample generation, $p_{s}(t)$ is fixed as $p_{s}^{-}$ for $t\in[0, 0.5)$ and as $p_{s}^{+}$ for $t\in[0.5, 1]$. Thus, the data generation process follows the settings of triplet $(p_{s}^{-}, p_{s}^{+}, p_{v})$. Evaluation on four test environments is performed with $p_{s}\in\{0.999, 0.8, 0.2, 0.001\}$ and $p_{v}$ unchanged. Time $t$  is used as the auxiliary variable in ZIN and \ref{eqn:MinimaxTVL1} for environment inference.

We present the mean accuracy and the worst accuracy over the four test environments in Table \ref{synthetic}. When environment partitions are unavailable, \ref{eqn:MinimaxTVL1} outperforms the other competitors in all but one case where $(p_{s}^{-}, p_{s}^{+})=(0.999, 0.7)$ and $p_{v}(t)=0.8$. When environment partitions are available, \ref{eqn:IRMv1wTVL1} is the best method. As the correlation between $X_{s}(t)$ and $Y(t)$ grows, the TV-$\ell_{1}$-based models are more robust in two aspects: 1. The gap between the mean and the worst performance of the TV-$\ell_{1}$-based models is generally smaller than that of the other competitors; 2. The performance of the TV-$\ell_{1}$-based models is less affected by such change.

We also show how the TV-$\ell_{1}$-based models improve identifiability of invariant features over the TV-$\ell_{2}$-based models in Appendix \ref{sec:identifiable}.

\begin{table*}[h]
\caption{Upper: accuracies (\%) of competing methods on four test environments in simulation study averaged by 10 repetitions. Lower:  STDs computed by 10 repetitions.}
\label{synthetic}
%\vskip 0.15in
\begin{center}
\begin{small}\scriptsize
\begin{sc}
\begin{tabular}{c|p{1.8cm}|p{0.6cm}|p{0.6cm}|p{0.6cm}|p{0.6cm}|p{0.6cm}|p{0.6cm}|p{0.6cm}|p{0.6cm}|p{0.6cm}|p{0.6cm}|p{0.6cm}|p{0.6cm}}
\toprule
%\abovespace\belowspace
\multirow{3}{*}{Env partition}    &  $(p_{s}^{-}, p_{s}^{+})$ & \multicolumn{4}{|c|}{(0.999, 0.7)}                        &\multicolumn{4}{|c|}{(0.999, 0.8)}                          &\multicolumn{4}{c}{(0.999, 0.9)}    \\
\cmidrule{2-6}\cmidrule {7-10}\cmidrule{11-14}
                    &  $p_{v}(t)$                   & \multicolumn{2}{|c|}{0.9}&\multicolumn{2}{|c|}{0.8} & \multicolumn{2}{|c|}{0.9} & \multicolumn{2}{|c|}{0.8} & \multicolumn{2}{|c|}{0.9} &\multicolumn{2}{c}{0.8}    \\
\cmidrule{2-4}\cmidrule{5-6}\cmidrule{7 -8}\cmidrule{9-10}\cmidrule{11-12}\cmidrule{13-14}\cmidrule {7-10}\cmidrule{11-14}
                    &   Method                &  Mean & Worst         & Mean & Worst       & Mean & Worst         &Mean & Worst         & Mean & Worst        &Mean & Worst  \\
\hline
\multirow{6}{*}{False} & ERM                        &   76.22 & 58.81 & 59.80 & 25.95 & 69.34 & 43.06 & 55.96 & 15.60 & 60.62 & 23.30 & 53.10 & 8.04 \\
                              &  EIIL                          &   39.43 & 18.22 & 64.95 & 48.45 & 50.26 & 47.02 & 68.86 & 54.91 & 61.33 & 52.70 & 69.82 & 58.58 \\
                              & HRM                        &  76.52 & 59.78 & 59.98 & 26.97 & 69.87 & 44.49 & 56.40 & 16.85 & 60.57 & 23.46 & 53.16 & 8.37 \\
                              & TIVA                        & 82.54 & 76.74 & 75.82 & 70.97 & 81.53 & 73.05 & 69.78 & 56.23 & 71.42 & 49.95 & 59.47 & 30.77 \\
                              & ZIN                          & 87.70 & 85.86 & \textbf{78.33} & 76.60 & 86.78 & 84.86 & 77.42 & 75.12 & 83.42 & 78.62 & 74.03 & 67.45 \\
                              & \textbf{Minmax-TV-$\ell_{1}$} & \textbf{88.67} & \textbf{87.83} & 78.14 & \textbf{76.68} & \textbf{88.55} & \textbf{87.62} & \textbf{78.74} & \textbf{77.56} & \textbf{87.01} & \textbf{85.74} & \textbf{77.31} & \textbf{74.54} \\
\hline
\multirow{3}{*}{True}  & groupDRO                 &  72.42 & 54.90 & 63.74 & 43.37 & 71.09 & 51.60 & 62.78 & 40.21 & 69.67 & 47.72 & 61.81 & 36.44 \\
                               & IRM                          & 87.84 & 86.20 & 78.33 & 76.58 & 86.84 & 84.42 & 77.48 & 74.80 & 84.16 & 77.89 & 74.53 & 68.72 \\
                               &\textbf{IRM-TV-$\ell_{1}$}      &  \textbf{88.03} & \textbf{86.40} & \textbf{78.49} & \textbf{76.88} & \textbf{87.10} & \textbf{84.90} & \textbf{77.95} & \textbf{75.65} & \textbf{84.84} & \textbf{80.06} & \textbf{75.55} & \textbf{70.77} \\
\bottomrule
\bottomrule
\multirow{6}{*}{False} & ERM                        &   1.17 & 2.06 & 1.04 & 2.06 & 1.23 & 2.47 & 0.76 & 1.42 & 1.10 & 2.01 & 0.62 & 0.95 \\
                              &  EIIL                          &   1.52 & 3.18 & 1.46 & 1.72 & 1.70 & 3.09 & 1.43 & 2.26 & 2.46 & 1.99 & 1.58 & 2.04 \\  
                              & HRM                        &  1.35 & 2.71 & 0.94 & 2.43 & 0.75 & 1.83 & 0.71 & 2.33 & 0.84 & 1.29 & 0.45 & 0.93 \\
                              & TIVA                        & 6.12 & 11.09 & 3.55 & 7.18 & 4.83 & 9.19 & 6.46 & 13.96 & 5.18 & 10.34 & 6.32 & 13.66 \\  
                              & ZIN                          & 1.05 & 2.19 & 1 & 1.43 & 1.67 & 2.73 & 1.43 & 2.13 & 3.52 & 6.72 & 2.09 & 3.86 \\
                              & \textbf{Minmax-TV-$\ell_{1}$} & 0.57 & 0.60 & 0.84 & 1.03 & 0.45 & 0.50 & 0.67 & 0.74 & 1.28 & 1.66 & 0.65 & 1.13 \\
\hline
\multirow{3}{*}{True}  & groupDRO                 &  8.45 & 18.08 & 6.99 & 16.84 & 8.42 & 19.03 & 6.71 & 17.27 & 8.27 & 18.51 & 6.52 & 16.45 \\ 
                               & IRM                          & 0.82 & 2.01 & 0.91 & 1.49 & 1.16 & 2.34 & 1.82 & 3.01 & 1.98 & 4.11 & 3.14 & 4.52 \\
                               &\textbf{IRM-TV-$\ell_{1}$}      &  0.86 & 2.08 & 0.74 & 1.33 & 1.35 & 2.67 & 1.24 & 2.22 & 2.19 & 4.77 & 2.92 & 4.31 \\
\bottomrule
\end{tabular}
\end{sc}
\end{small}
\end{center}
\vspace{-15pt}
\end{table*}

\subsection{Real-world Experiments}
{\bf{House Price Prediction}}. We use the House Prices data set\footnote{https://www.kaggle.com/c/house-prices-advanced-regression-techniques/data} to verify the TV-$\ell_{1}$-based models in a regression task. In this experiment, $15$ variables including the number of bathrooms, locations, etc., are used to predict the house price. Samples with built year in period $[1900, 1950]$ are used for training and those with built year in period $(1950, 2000]$ are used for test. The house price is normalized within the same built year. The built year is used as an auxiliary variable in both ZIN and \ref{eqn:MinimaxTVL1} for environment inference. The training samples are divided into $5$ segments with $10$-year range in each segment. Then each segment is considered as having the same environment.

Table \ref{houseprice} reports the mean squared errors (MSE) of competing methods in this regression task. \ref{eqn:IRMv1wTVL1} achieves the best results in the environment-partition case, and \ref{eqn:MinimaxTVL1} achieves the best results in the non-environment-partition case. Again, the TV-$\ell_{1}$-based models have smaller gaps between the mean and the worst performance than the other competitors.

\begin{table}[t]
\caption{Upper: mean squared errors of competing methods in house price prediction averaged by 10 repetitions. Lower:  STDs computed by 10 repetitions. }
\label{houseprice}
\vskip 0.15in
\begin{center}
\begin{small}\scriptsize
\begin{sc}
\begin{tabular}{p{1.7cm}p{1.8cm}p{0.6cm}p{0.6cm}p{0.6cm}p{0.6cm}}
\toprule
%\abovespace\belowspace
Env Partition & Method & Train & Test & Worst \\
\hline
%\abovespace
\multirow{6}{*}{False}  & ERM 		&	0.1057 & 0.4409 & 0.6206 \\
                                & EIIL   	&  0.1103 & 0.3939 & 0.5581  \\
                                & HRM 				& 0.5578 & 0.5949 & 0.7250  \\
                                & TIVA 				& 0.2575 & 0.4418 & 0.6145  \\
                                & ZIN				& 0.2241 & 0.4293 & 0.6198  \\
                                &\textbf{Minmax-TV-$\ell_{1}$}& 0.2168 & \textbf{0.3395} & \textbf{0.4983}  \\
\hline
\multirow{3}{*}{True} & groupDRO 			& 0.1271 & 0.7358 & 1.0611  \\
				 & IRM				& 0.5663 & 0.8168 & 1.1168  \\
				 & \textbf{IRM-TV-$\ell_{1}$} 	& 0.3261 & \textbf{0.4420} & \textbf{0.6096}  \\
\bottomrule
\bottomrule
\multirow{6}{*}{False}  & ERM 		 & 0.0017 & 0.0435 & 0.0641 \\
                                & EIIL   	 & 0.0020 & 0.0305 & 0.0460 \\
                                & HRM 				 & 0.0593 & 0.0025 & 0.0052 \\
                                & TIVA 				 & 0.0002 & 0.0019 & 0.0062 \\
                                & ZIN				 & 0.1137 & 0.1994 & 0.2869 \\
                                &\textbf{Minmax-TV-$\ell_{1}$} & 0.0652 & 0.0638 & 0.0958 \\
\hline
\multirow{3}{*}{True} & groupDRO 			 & 0.0029 & 0.0877 & 0.1287 \\
				 & IRM				 & 0.1389 & 0.3115 & 0.4511 \\
				 & \textbf{IRM-TV-$\ell_{1}$} 	 & 0.1279 & 0.2503 & 0.3342 \\
\bottomrule
\end{tabular}
\end{sc}
\end{small}
\end{center}
\vspace{-12pt}
\end{table}

{\bf{CelebA}}. This data set contains face images of celebrities \cite{celebA}. The task is to identify the smiling faces, which is deliberately correlated with gender. $512$-dimensional deep features of face images are extracted by a pre-trained ResNet18 \cite{resnet}, and the invariant features are learned by subsequent multilayer perceptrons. Seven additional descriptive variables including \emph{Young}, \emph{Blond Hair}, \emph{Eyeglasses}, \emph{High Cheekbones}, \emph{Big Nose}, \emph{Bags Under Eyes}, and \emph{Chubby} are fed into ZIN and \ref{eqn:MinimaxTVL1} for environment inference. The gender variable is only used as the environment indicator for groupDRO, IRM, and \ref{eqn:IRMv1wTVL1}. 

Table \ref{celeba} presents the results. The TV-$\ell_{1}$-based models achieve the best accuracies in the mean and the worst scenarios whether environment partitions are available or not. Moreover, \ref{eqn:MinimaxTVL1} is closer to IRM and \ref{eqn:IRMv1wTVL1} than other environment inference methods (EIIL, LfF, ZIN, and TIVA). It indicates that TV-$\ell_{1}$ narrows the gap between absence and presence of environment information.

\begin{table}[t]
\caption{Upper: accuracies (\%) of competing methods on CelebA averaged by 10 repetitions. Lower:  STDs computed by 10 repetitions.}
\label{celeba}
\vskip 0.15in
\begin{center}
\begin{small}\scriptsize
\begin{sc}
\begin{tabular}{p{1.7cm}p{1.8cm}p{0.6cm}p{0.6cm}p{0.6cm}p{0.6cm}}
\toprule
%\abovespace\belowspace
Env Partition & Method & Train & Test & Worst \\
\hline
%\abovespace
\multirow{6}{*}{False}  & ERM 				& 63.76 & 63.99 & 62.05  \\
                                & EIIL   	&  59.12 & 58.15 & 54.22  \\
                                & LfF 		& 57.50 & 57.73 & 56.18  \\
                                & TIVA 		& 64.36 & 64.23 & 61.63  \\
                                & ZIN		& 78.32 & 76.73 & 76.19  \\
                                &\textbf{Minmax-TV-$\ell_{1}$}& 85.12 & \textbf{83.68} & \textbf{81.45}  \\
\hline
\multirow{3}{*}{True} & groupDRO 			& 81.50 & 81.19 & 79.27  \\
				 & IRM				& 85.59 & 82.54 & 80.75  \\
				 & \textbf{IRM-TV-$\ell_{1}$} 	& 84.79 & \textbf{83.47} & \textbf{81.21}  \\
\bottomrule
\bottomrule
\multirow{6}{*}{False}  & ERM 				 & 14.45 & 14.16 & 14.16 \\
                                & EIIL   	 & 8.74 & 8.48 & 10.23 \\
                                & LfF 		 & 0.12 & 0.24 & 0.57 \\
                                & TIVA 		 & 1.68 & 1.99 & 1.47 \\
                                & ZIN		 & 1.16 & 0.87 & 0.85 \\
                                &\textbf{Minmax-TV-$\ell_{1}$} & 0.92 & 0.33 & 0.43 \\
\hline
\multirow{3}{*}{True} & groupDRO 			 & 0.31 & 0.48 & 0.74 \\
				 & IRM				 & 1.49 & 1.35 & 0.99 \\
				 & \textbf{IRM-TV-$\ell_{1}$} 	 & 0.59 & 0.48 & 0.67 \\
\bottomrule
\end{tabular}
\end{sc}
\end{small}
\end{center}
\vskip -0.1in
\end{table}

{\bf{Landcover}}. The Landcover data set records time series and the corresponding land cover types from the satellite data \cite{landcover2006, landcover2020, landcover2021}. Time series data with dimension $46\times 8$ are used as input to identify one from six land cover types. The invariant feature extractor $\Phi$ is instantiated as a 1D-CNN to handle the time series input, following \cite{landcover2021, ZIN}. Ground-truth environment partitions are unavailable, thus latitude and longitude are taken as auxiliary information for environment inference. All methods are trained on non-African data, and then tested on both non-African (not overlapping the training data) and African data. Corresponding results are denoted by IID Test and OOD Test.

As shown in Table \ref{landcover}, \ref{eqn:MinimaxTVL1} achieves the best performance in all of the IID, OOD and Worst Tests, with at least $1.4\%$ higher than the second best competitor. Hence \ref{eqn:MinimaxTVL1} can be better generalized to unseen environments.

\begin{table}[t]
\caption{Upper: accuracies (\%) of competing methods on Landcover averaged by 10 repetitions. Lower:  STDs computed by 10 repetitions.}
\label{landcover} 
\vskip 0.15in
\begin{center}
\begin{small}%\scriptsize
\begin{sc}
\scalebox{0.9}{\begin{tabular}{ccccc}
\toprule
%\abovespace\belowspace
 Method & Train & IID Test & OOD Test  & Worst\\
 \hline
 ERM	   &       66.61 & 66.44 & 61.54 & 60.80  \\
 EIIL		 &      64.11 & 63.81 & 60.43 & 59.53  \\
 LfF		 &      58.12 & 57.89 & 55.76 & 55.07  \\
 TIVA	    &       67.49 & 64.79 & 52.02 & 51.46  \\
 ZIN 		 & 	 70.02 & 69.42 & 62.22 & 61.87  \\
 \textbf{Minmax-TV-$\ell_{1}$} &  73.59 & \textbf{71.95} & \textbf{63.77} & \textbf{63.25}  \\
\bottomrule
\bottomrule
 ERM	    & 1.82 & 1.56 & 0.92 & 0.77 \\
 EIIL		  & 1.66 & 1.72 & 0.88 & 1.21 \\
 LfF		  & 2.73 & 2.45 & 1.96 & 1.93 \\
 TIVA	     & 0.28 & 0.62 & 0.98 & 1.09 \\
 ZIN 		  & 1.09 & 1.14 & 1.09 & 1.21 \\
 \textbf{Minmax-TV-$\ell_{1}$}  & 0.69 & 0.63 & 1.17 & 1.37 \\
\bottomrule
\end{tabular}}
\end{sc}
\end{small}
\end{center}
\vskip -0.3in
\end{table}

{\bf Adult Income Prediction}. In this task we use the Adult data set\footnote{https://archive.ics.uci.edu/dataset/2/adult} to predict if the income of an individual exceeds \$50K/yr based on the census data. We split the data set into four subgroups  regarded as separated environments according to $race \in \{\text{Black}, \text{Non-Black}\}$ and $sex \in \{\text{Male}, \text{Female}\}$. We randomly choose two thirds of data from the subgroups Black Male and Non-Black Female for training, and then verify models across all four subgroups with the rest data. Six integral variables: \emph{Age}, \emph{FNLWGT}, \emph{Eduction-Number}, \emph{Capital-Gain}, \emph{Capital-Loss}, and \emph{Hours-Per-Week} are fead into ZIN and \ref{eqn:MinimaxTVL1} for environment inference. Ground-truth environment indicators are provided for groupDRO, IRM and \ref{eqn:IRMv1wTVL1}. Categorical variables except race and sex are encoded by one-hot coding, followed by the principal component analysis transform with over 99\% cumulative explained variance ratio kept. The transformed features are combined with the integral variables, yielding 59-dimensional representations, which are subsequently normalized to have zero mean and unit variance for invariant feature learning. 

Results are shown in Table \ref{adult}. \ref{eqn:MinimaxTVL1} and \ref{eqn:IRMv1wTVL1} achieve the best accuracies in the mean and the worst scenarios within their respective categories. Again, \ref{eqn:MinimaxTVL1} is closer to IRM and \ref{eqn:IRMv1wTVL1} than other environment inference methods. Hence \ref{eqn:MinimaxTVL1} narrows the gap between absence and presence of environment information.

\vspace{-5pt}
\begin{table}[h]
\caption{Upper: accuracies (\%) of competing methods in adult income prediction averaged by 10 repetitions. Lower:  STDs computed by 10 repetitions. }
\label{adult}
%\vskip 0.15in
\begin{center}
\begin{small}\scriptsize
\begin{sc}
\begin{tabular}{p{1.7cm}p{1.8cm}p{0.6cm}p{0.6cm}p{0.6cm}}
\toprule
Env Partition & Method & Train & Test & Worst   \\
\hline
%\abovespace
\multirow{6}{*}{False}  & ERM 				& 93.34 & 82.16 & 79.55  \\
                                & EIIL   	& 79.97 & 72.77 & 70.94  \\
                                & LfF 		& 82.03 & 75.04 & 73.00  \\
                                & TIVA 		& 91.45 & 81.95 & 79.28  \\
                                & ZIN		& 93.16 & 82.26 & 79.67 \\
                                &\textbf{Minmax-TV-$\ell_{1}$} & 92.40 & \textbf{83.33} & \textbf{80.95}  \\
\hline
\multirow{3}{*}{True} & groupDRO 			& 87.51 & 76.42 & 73.07 \\
				 & IRM				& 93.19 & 82.32 & 79.76  \\
				 & \textbf{IRM-TV-$\ell_{1}$} 	& 92.42 & \textbf{83.31} & \textbf{80.93} \\
\bottomrule
\bottomrule
\multirow{6}{*}{False}  & ERM 				 & 0.31 & 0.33 & 0.37 \\
                                & EIIL   	 & 0.56 & 0.61 & 0.73 \\
                                & LfF 		 & 5.54 & 3.01 & 2.45 \\
                                & TIVA 		 & 0.12 & 0.39 & 0.46 \\
                                & ZIN		 & 0.17 & 0.27 & 0.29 \\
                                &\textbf{Minmax-TV-$\ell_{1}$}  & 0.11 & 0.14 & 0.16 \\
\hline
\multirow{3}{*}{True} & groupDRO 			 & 0.59 & 1.29 & 1.54 \\
				 & IRM				 & 0.28 & 0.23 & 0.29 \\
				 & \textbf{IRM-TV-$\ell_{1}$} 	 & 0.19 & 0.18 & 0.19 \\
\bottomrule
\end{tabular}
\end{sc}
\end{small}
\end{center}
%\vskip -0.1in
\end{table}

\vspace{-5pt}

\section{Conclusion and Discussion}
We theoretically show that IRM is essentially an IRM-TV-$\ell_{2}$ model that simultaneously minimizes the total empirical risk and its total variation with respect to the classifier variable. Following this idea, we propose the IRM-TV-$\ell_{1}$ and the Minimax-TV-$\ell_1$ models for learning tasks with or without environment partitions, respectively. They allow broader classes of functions to be the learning risk and the feature extractor, and preserve invariant features based on the coarea formula. Moreover, we investigate the requirements for the TV-$\ell_1$ framework to achieve out-of-distribution (OOD) generalization. Extensive experiments on both synthetic and real-world data sets show that the TV-$\ell_1$ framework performs well in several OOD tests. 

A direct impact of this work is to provide a new technical approach that identifies, analyzes and constructs different kinds of invariants with nonsmooth or even discontinuous modules. Although these properties are difficult to handle, they precisely show a reasonable nature of generalization ability and robustness. Further improvements on this work may lie in designing an adaptive penalty parameter to improve OOD generalization, constructing a diverse and representative training environment space, and developing new TV-$\ell_1$ models for deep learning. These are nontrivial tasks and we shall put them in future works.

\section*{Acknowledgments}
We would like to thank the anonymous reviewers for their constructive suggestions that help to improve this paper.

This work is supported in part by the National Natural Science Foundation of China under grant 62176103, in part by the Science and Technology Planning Project of Guangzhou under grants 2024A04J9896, 2024A04J4225, and in part by the Fundamental Research Funds for the Central Universities under grant 21623341.

Code is available at \url{https://github.com/laizhr/IRM-TV}.

\section*{Impact Statement}
This paper presents work whose goal is to advance the field of Machine Learning. There are no potential societal consequences of our work that we feel must be specifically highlighted here.

\bibliography{bibfile}
\bibliographystyle{icml2024}

%%%%%%%%%%%%%%%%%%%%%%%%%%%%%%%%%%%%%%%%%%%%%%%%%%%%%%%%%%%%%%%%%%%%%%%%%%%%%%%
%%%%%%%%%%%%%%%%%%%%%%%%%%%%%%%%%%%%%%%%%%%%%%%%%%%%%%%%%%%%%%%%%%%%%%%%%%%%%%%
% APPENDIX
%%%%%%%%%%%%%%%%%%%%%%%%%%%%%%%%%%%%%%%%%%%%%%%%%%%%%%%%%%%%%%%%%%%%%%%%%%%%%%%
%%%%%%%%%%%%%%%%%%%%%%%%%%%%%%%%%%%%%%%%%%%%%%%%%%%%%%%%%%%%%%%%%%%%%%%%%%%%%%%
\newpage
\appendix
\onecolumn
\setcounter{table}{0}
\renewcommand{\thetable}{A\arabic{table}}

\setcounter{figure}{0}
\renewcommand{\thefigure}{A\arabic{figure}}

\section{Proofs}

\subsection{Proof of Theorem \ref{thm:IRMTVL2}}
\label{proof:thmIRMTVL2}

\subsubsection{Induced Measure $\nu$}
\label{proof:thmIRMTVL2inducemeasure}
%\label{proof:thmIRMTVL2}
First, we verify that there is a measure $\nu$ for $w$ induced by $\mu$. From Condition 3, let $\Omega=w(\mE_{tr})$ and $\sF_{\Omega}:=\{w(\mF): \mF\in\sF_{tr}\}$. By definition, $\sF_{\Omega}$ is the $\sigma$-algebra for $\Omega$ under mapping $w$. Define a set function $\nu$ as follows:
\begin{align}
\label{eqn:nudef}
\nu: \sF_{\Omega}\rightarrow \bbR,\quad \nu(\mG):=\mu(w^{-1}(\mG)).
\end{align}
It is well defined because $w^{-1}(\mG)\in \sF_{tr}$. Now we prove that $\nu$ is a measure. Since $\mu$ is a measure, the following non-negativity and zero measure are obvious:
\begin{align}
\label{eqn:measnonneg}
\forall \ \mG\in \sF_{\Omega}, \quad\nu(\mG)=\mu(w^{-1}(\mG))\geqs 0, \quad\nu(\emptyset)=\mu(w^{-1}(\emptyset))=\mu(\emptyset)=0.
\end{align}
To prove countable additivity, let $\{\mG_i\}_{i=1}^{\infty}$ be any countable collection of mutually disjoint sets in $\sF_{\Omega}$. From the property of measurable function, $\{w^{-1}(\mG_i)\}_{i=1}^{\infty}$ is also a countable collection of mutually disjoint sets in $\sF_{tr}$. Therefore,
\begin{align}
\label{eqn:meascnt}
\nu(\bigcup_{i=1}^{\infty}\mG_i)=\mu(w^{-1}(\bigcup_{i=1}^{\infty}\mG_i))=\mu(\bigcup_{i=1}^{\infty}[w^{-1}(\mG_i)])=\sum_{i=1}^{\infty}\mu(w^{-1}(\mG_i))=\sum_{i=1}^{\infty}\nu(\mG_i),
\end{align}
where the mutually disjoint $\{\mG_i\}_{i=1}^{\infty}$ yields the second equality and the mutually disjoint $\{w^{-1}(\mG_i)\}_{i=1}^{\infty}$ yields the third equality, respectively. Combining \eqref{eqn:measnonneg} and \eqref{eqn:meascnt}, we can conclude that $\nu$ is a measure.

\subsubsection{Constructing Measurable Functions $R(w\circ \Phi)$ and $|\nabla_w R(w\circ \Phi)|$}
\label{proof:thmIRMTVL2measurefunc}
From Conditions 2 and 4, $R(w(e)\circ \Phi): \mE_{tr}\rightarrow \bbR$ is a measurable function on $(\mE_{tr},\sF_{tr})$ and
\begin{align}
\label{eqn:intchange}
\int_{\mE_{tr}} R(w\circ \Phi,e)\dmu=\int_{\mE_{tr}} R(w(e)\circ \Phi)\dmu\in (-\infty,+\infty).
\end{align}
We can directly construct a function $R(w\circ \Phi): \Omega\rightarrow \bbR$ on $(\Omega,\sF_{\Omega})$ and verify its measurability:
\begin{align}
\label{eqn:directdefRw}
\forall \omega\in \Omega,\quad R(\omega\circ \Phi):=R(w[w^{-1}(\omega)]\circ \Phi),
\end{align}
where $w^{-1}(\omega)\subseteq\mE_{tr}$. From Condition 3, we have $\mE=w^{-1}(\mG)\in \sF_{tr}$ for any $\mG\in \sF_{\Omega}$. Hence 
\begin{align}
\label{eqn:directdefRwsigalg}
R(\mG\circ \Phi): \sF_{\Omega}\rightarrow \bbR,  \quad R(\mG\circ \Phi):=R(w[w^{-1}(\mG)]\circ \Phi)=R(w(\mE)\circ \Phi)
\end{align}
is well-defined. Next, let $\mR_0\in \sF_{\bbR}$, where $\sF_{\bbR}$ denotes the $\sigma$-algebra of all the Lebesgue measurable sets contained in $\bbR$. Then there exists some $\mE_0\in \sF_{tr}$ such that 
\begin{align}
\label{eqn:directdefmeasurable}
R(w(\mE_0)\circ \Phi)=\mR_0.
\end{align}
Let $\mG_0:=w(\mE_0)\in \sF_{\Omega}$. Since $\mR_0$ is arbitrary, we have
\begin{align}
\label{eqn:directdefmeasurable2}
\mathring{\sF}_{\Omega}:=\{\mG_0:\ R(\mG_0\circ \Phi)=\mR_0,   \ \mR_0 \in \sF_{\bbR}\}\subseteq\sF_{\Omega}.
\end{align}

It remains to verify that $\mathring{\sF}_{\Omega}$ is a $\sigma$-algebra, forming a sub-$\sigma$-algebra of $\sF_{\Omega}$. It is evident that $\emptyset, \Omega\in \mathring{\sF}_{\Omega}$. For any $\mG_0\in \mathring{\sF}_{\Omega}$,
\begin{align}
\label{eqn:directdefmeasurable3}
\Omega \backslash \mG_0=w[w^{-1}(\Omega \backslash \mG_0)]=w[w^{-1}(\Omega )\backslash w^{-1}(\mG_0)]=w[\mE_{tr}\backslash w^{-1}(\mG_0)].
\end{align}
Because $\mG_0\in \mathring{\sF}_{\Omega}\subseteq\sF_{\Omega}$, $w^{-1}(\mG_0)\in \sF_{tr}$. Since $\sF_{tr}$ is closed under complement, $\mE_{tr}\backslash w^{-1}(\mG_0)\in \sF_{tr}$. Thus $w[\mE_{tr}\backslash w^{-1}(\mG_0)]\in {\sF}_{\Omega}$. Then
\begin{align}
\label{eqn:directdefmeasurable4}
R((\Omega \backslash \mG_0)\circ \Phi)=R(w[\mE_{tr}\backslash w^{-1}(\mG_0)]\circ \Phi)=\tilde{\mR}_0 \ \text{for some} \ \tilde{\mR}_0\in \sF_{\bbR}.
\end{align}
The first equality follows from \eqref{eqn:directdefmeasurable3} and the second equality follows from the measurability of $R(w(e)\circ \Phi)$ w.r.t. $e$. \eqref{eqn:directdefmeasurable4} indicates that $\Omega \backslash \mG_0\in \mathring{\sF}_{\Omega}$ according to \eqref{eqn:directdefmeasurable2}. Hence $\mathring{\sF}_{\Omega}$ is closed under complement. Finally, let any $\{\mG_i\in \mathring{\sF}_{\Omega}\}_{i=1}^\infty$. Then
\begin{align}
\label{eqn:directdefmeasurable5}
\bigcup_{i=1}^\infty \mG_i=w\left [w^{-1}(\bigcup_{i=1}^\infty \mG_i)\right ]=w\left [\bigcup_{i=1}^\infty w^{-1}(\mG_i)\right ].
\end{align}
Because $\mG_i\in \mathring{\sF}_{\Omega}\subseteq\sF_{\Omega}$, $w^{-1}(\mG_i)\in \sF_{tr}$. Since $\sF_{tr}$ is closed under countable unions, $\bigcup_{i=1}^\infty w^{-1}(\mG_i)\in \sF_{tr}$. Thus $w\left [\bigcup_{i=1}^\infty w^{-1}(\mG_i)\right ]\in {\sF}_{\Omega}$. Then 
\begin{align}
\label{eqn:directdefmeasurable6}
R\left (\left (\bigcup_{i=1}^\infty \mG_i\right )\circ \Phi \right )=R\left (w\left [\bigcup_{i=1}^\infty w^{-1}(\mG_i)\right ]\circ \Phi\right )=\mR_{\infty} \ \text{for some} \ \mR_{\infty}\in \sF_{\bbR}.
\end{align}
The first equality follows from \eqref{eqn:directdefmeasurable5} and the second equality follows again from the measurability of $R(w(e)\circ \Phi)$ w.r.t. $e$. \eqref{eqn:directdefmeasurable6} indicates that $\bigcup_{i=1}^\infty \mG_i\in \mathring{\sF}_{\Omega}$ according to \eqref{eqn:directdefmeasurable2}. Hence $\mathring{\sF}_{\Omega}$ is closed under countable unions. To summarize, $\mathring{\sF}_{\Omega}$ satisfies the definition of $\sigma$-algebra and forms a sub-$\sigma$-algebra of $\sF_{\Omega}$.

On the other hand, the measurable function $|\nabla_w R(w\circ \Phi)|$ can be constructed via the same procedure. Condition 2 implies that $\nabla_w R(w\circ \Phi,e)\dmu\equiv \nabla_w R(w(e)\circ \Phi)\dmu$ since the partial differential is taken on $w$. Moreover, Condition 4 indicates that
\begin{align}
\label{eqn:intchange2}
\int_{\mE_{tr}} |\nabla_w R(w\circ \Phi,e)|^2\dmu=\int_{\mE_{tr}} |\nabla_w R(w(e)\circ \Phi)|^2\dmu\in (-\infty,+\infty).
\end{align}
We are now ready to establish the integrations w.r.t. $w$ from \eqref{eqn:intchange} and \eqref{eqn:intchange2}.

\subsubsection{Integrations w.r.t. $w$}
\label{proof:thmIRMTVL2integral}
The integrations w.r.t. $w$ can be established via the integrations w.r.t. $e$. First, we aim to establish the following:
\begin{align}
\label{eqn:intchangew}
\int_{\Omega} R(w\circ \Phi) \dnu :=\int_{\mE_{tr}} R(w(e)\circ \Phi)\dmu, \quad \forall \ R(w(e)\circ \Phi)\in L^1(\mE_{tr},\sF_{tr},\mu).
\end{align}
\ref{proof:thmIRMTVL2inducemeasure} and \ref{proof:thmIRMTVL2measurefunc} have already verified the eligible measure $\dnu$ and the measurable function $R(w\circ \Phi)$. We start from the simplest case. Let $\mG$ be any measurable set in $\sF_{\Omega}$ and $\mE=w^{-1}(\mG)\in\sF_{tr}$. Define the indicator function of $\mG$ as follows:
\begin{align}
\label{eqn:intecator}
\bbI_{\mG}(w)= \left\{\begin{array}{ll}
1 & \text{if}\quad w\in \mG,\\
0 & \text{if}\quad w\notin \mG.
\end{array} \right.
\end{align}
We have
\begin{align}
\label{eqn:intchangewsimple}
\int_{\Omega} \bbI_{\mG}(w) \dnu=\nu(\mG)=\mu(\mE)=\int_{\mE_{tr}}\bbI_{\mE}(e) \dmu.
\end{align}
This relationship also holds for $a\bbI_{\mG}(w)$, $\forall a\geqs 0$. Next, define the sets of non-negative simple functions as
\begin{align}
\label{eqn:nonnegsimp}
\mS^+_{\Omega}:=\left \{\sum_{i=1}^m a_i\bbI_{\mG_i}(w): m\in \bbN^+, a_i\geqs 0, \mG_i\in\sF_{\Omega}  \right \}\quad \text{and}\quad  \mS^+_{tr}:=\left \{\sum_{i=1}^m a_i\bbI_{\mE_i}(e): m\in \bbN^+, a_i\geqs 0, \mE_i\in\sF_{tr}  \right \}.
\end{align}
Then for any $\sum_{i=1}^m a_i\bbI_{\mG_i}(w)$, denote $\mE_i=w^{-1}(\mG_i)$, $i=1,\cdots,m$. We have 
\begin{align}
\label{eqn:intchangewstair}
\int_{\Omega}\sum_{i=1}^m a_i \bbI_{\mG_i}(w) \dnu =\sum_{i=1}^m a_i\int_{\Omega} \bbI_{\mG_i}(w) \dnu=\sum_{i=1}^m a_i\nu(\mG_i)=\sum_{i=1}^m a_i\mu(\mE_i)=\sum_{i=1}^m a_i\int_{\mE_{tr}}\bbI_{\mE_i}(e) \dmu=\int_{\mE_{tr}}\sum_{i=1}^m a_i\bbI_{\mE_i}(e) \dmu.
\end{align}
That is, for any $\sum_{i=1}^m a_i \bbI_{\mG_i}(w)\in\mS^+_{\Omega}$, there exists some $\sum_{i=1}^m a_i\bbI_{\mE_i}(e)\in\mS^+_{tr}$ satisfying \eqref{eqn:intchangewstair}.

Next, suppose $R^+(w\circ \Phi)$ is a \textbf{non-negative measurable function} w.r.t. $w$. Then it can be approached by a sequence of functions in $\mS^+_{\Omega}$. Specifically, let
\begin{align}
\label{eqn:intchangewstair2}
a_i:=\frac{i}{2^n}, \mG_i:=\{w: a_i\leqs R(w\circ \Phi)< a_{i+1}\},\ i=0,1,\cdots,n2^n-1; \ \mG_{R\geqs n}:=\{w: R(w\circ \Phi)\geqs n\}.
\end{align}
Define the sequence of approaching functions as follows:
\begin{align}
\label{eqn:funcapprachw}
R^+_n(w\circ \Phi)=\sum_{i=0}^{n2^n-1} a_i\bbI_{\mG_i}(w)+n\bbI_{\mG_{R\geqs n}}(w),\quad n=1,2,3,\cdots. 
\end{align}
It can be easily observed that $R^+_n(w\circ \Phi)\in\mS^+_{\Omega}$ and $R^+_n(w\circ \Phi)\uparrow R^+(w\circ \Phi)$ (point-wise monotonically non-decreasing and convergent). Let $\mE_i=w^{-1}(\mG_i)$, $i=0,1,\cdots,n2^n-1$, and $\mE_{R\geqs n}=w^{-1}(\mG_{R\geqs n})$. Then 
\begin{align}
\label{eqn:funcapprache}
R^+_n(w(e)\circ \Phi)=\sum_{i=0}^{n2^n-1} a_i\bbI_{\mE_i}(e)+n\bbI_{\mE_{R\geqs n}}(e),\quad n=1,2,3,\cdots. 
\end{align}
We also have $R^+_n(w(e)\circ \Phi)\in\mS^+_{tr}$ and $R^+_n(w(e)\circ \Phi)\uparrow R^+(w(e)\circ \Phi)$ for some non-negative measurable function $R^+(w(e)\circ \Phi)$ w.r.t. $e$. From \eqref{eqn:intchangewstair},
\begin{align}
\label{eqn:intchangewstairseq}
\int_{\Omega} R^+_n(w\circ \Phi) \dnu =\int_{\mE_{tr}} R^+_n(w(e)\circ \Phi)\dmu,\quad n=1,2,3,\cdots.
\end{align}
Moreover, this sequence of integrals is monotonically increasing and thus converges to $\int_{\mE_{tr}} R^+(w(e)\circ \Phi)\dmu\in \bar{\bbR}$ (the extended real space including $+\infty$ and $-\infty$). We can directly define 
\begin{align}
\label{eqn:intchangewnonnegmeas}
\int_{\Omega} R^+(w\circ \Phi) \dnu :=\int_{\mE_{tr}} R^+(w(e)\circ \Phi)\dmu.
\end{align}
Since we only consider integrable functions $R^+(w(e)\circ \Phi)\in L^1(\mE_{tr},\sF_{tr},\mu)$ by Condition 4, \eqref{eqn:intchangewnonnegmeas} indicates that $\int_{\Omega} R^+(w\circ \Phi) \dnu<+\infty$.

Last, suppose $R(w\circ \Phi)$ is a \textbf{measurable function} w.r.t. $w$. Following conventional methodology, it can be decomposed into a positive part and a negative part: 
\begin{align}
\label{eqn:measdecom}
R(w\circ \Phi)=R^+(w\circ \Phi)-R^-(w\circ \Phi),\quad R^+(w\circ \Phi):=\max\{R(w\circ \Phi),0\},\ R^-(w\circ \Phi):=\max\{-R(w\circ \Phi),0\}.
\end{align}
Both $R^+(w\circ \Phi)$ and $R^-(w\circ \Phi)$ are non-negative measurable functions. By \eqref{eqn:intchangewnonnegmeas}, there exist non-negative measurable functions $R^+(w(e)\circ \Phi)$ and $R^-(w(e)\circ \Phi)$ w.r.t. $e$ such that
\begin{align}
\label{eqn:intchangewnonnegmeas2}
\int_{\Omega} R^+(w\circ \Phi) \dnu :=\int_{\mE_{tr}} R^+(w(e)\circ \Phi)\dmu, \quad\int_{\Omega} R^-(w\circ \Phi) \dnu :=\int_{\mE_{tr}} R^-(w(e)\circ \Phi)\dmu.
\end{align}
Let $R(w(e)\circ \Phi)=R^+(w(e)\circ \Phi)-R^-(w(e)\circ \Phi)$. Again, we only consider $R(w(e)\circ \Phi)\in L^1(\mE_{tr},\sF_{tr},\mu)$ by Condition 4, thus both integrals in \eqref{eqn:intchangewnonnegmeas2} are finite. Define 
\begin{align}
\label{eqn:intchangewmeas}
\int_{\Omega} R(w\circ \Phi) \dnu:=\int_{\Omega} R^+(w\circ \Phi) \dnu -\int_{\Omega} R^-(w\circ \Phi) \dnu,
\end{align}
then $\int_{\Omega} R(w\circ \Phi) \dnu\in (-\infty,+\infty)$.

One can apply the above procedure to $|\nabla_w R(w\circ \Phi)|^2$ and exploit $|\nabla_w R(w\circ \Phi)|\in  L^2(\mE_{tr},\sF_{tr},\mu)$ from Condition 4 to define
\begin{align}
\label{eqn:intchangewmeasnab}
\int_{\Omega}   |\nabla_w R(w\circ \Phi)|^2 \dnu\in [0,+\infty).
\end{align}
Note that $|\nabla_w R(w\circ \Phi)|$ is a non-negative measurable function.

\subsubsection{Expectations w.r.t. $w$ and Well-defined \eqref{eqn:IRMv1wTVL2}}
To ensure that $\int_{\Omega} R(w\circ \Phi) \dnu$ and $\int_{\Omega}   |\nabla_w R(w\circ \Phi)|^2 \dnu$ are mathematical expectations, we need to verify that $\nu$ is also a probability measure induced by $\mu$:
\begin{align}
\label{eqn:probinduced}
\nu(\Omega)=\mu(\mE_{tr})=1,
\end{align}
where $\mu(\mE_{tr})=1$ because $\mu$ is a probability measure. $\bbE_{w} [R(w\circ \Phi)]$ and $ \bbE_{w}   [|\nabla_w R(w\circ \Phi)|^2]$ denote the two integrals when $\nu$ is a probability measure. Since both integrals are finite, $\bbE_{w} [R(w\circ \Phi) +\lambda  |\nabla_w R(w\circ \Phi)|^2 ]$ is also finite and can be minimized w.r.t. $\Phi$. Hence \eqref{eqn:IRMv1wTVL2} is well-defined.

\textbf{For an extension, one can establish a probability measure as long as} $\nu(\Omega)\in (0,+\infty)$:
\begin{align}
\label{eqn:probesta}
\tilde{\nu}(\cdot):=\nu(\cdot)/\nu(\Omega).
\end{align}
\textbf{It is very useful when $\Omega$ is a bounded subset of $\bbRd$ and $\nu$ is the Lebesgue measure.}

\subsection{Proof of Proposition \ref{cor:VREx}}
\label{proof:VREx}
Given Conditions 1 and 4a, $\bbE_{e} [R(\Phi,e)]$ and $\bbE_{e} [R^2(\Phi,e)]$ are well-defined and finite. We can use the integral with probability measure $\mu$ to represent the first term in \eqref{eqn:VREx}:
\begin{align}
\label{eqn:sumRe}
\sum_{e\in \mE_{tr}}   R(\Phi,e):=\bbE_{e} [R(\Phi,e)]\in (-\infty,+\infty).
\end{align}
Additionally, the variance of $R(\Phi,e)$ in \eqref{eqn:VREx} can be generalized to
\begin{align}
\label{eqn:varRe}
\bbV (\{R(\Phi,e)\}_{e\in \mE_{tr}}):=\bbE_{e} [(R(\Phi,e){-}\bbE_{e} [R(\Phi,e)])^2]=\bbE_{e} [R^2(\Phi,e)]-(\bbE_{e} [R(\Phi,e)])^2\in (-\infty,+\infty).
\end{align}
Therefore, \eqref{eqn:VRExTVL2} is well-defined and finite, and can be minimized. 

\begin{figure}[!htb]
\centering
\includegraphics[width=0.6\columnwidth]{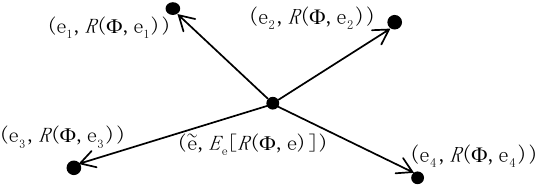}
\caption{Risk extrapolation of V-REx-$\ell_2$ to a larger region.}
\label{fig:riskextra}
\end{figure}

To illustrate how V-REx-$\ell_2$ extrapolates risk to a larger region, we need to examine the graph space of the learning risk: $\mE_{tr}\times \bbR:=\{(e,R(\Phi,e)): e\in \mE_{tr}\}$. $\bbE_{e} [R(\Phi,e)]$ can be seen as the mean risk of all the environments, which is located in the center. Under some mild assumptions (such as the continuity of $R(\Phi,e)$ w.r.t. $e$), $\bbE_{e} [R(\Phi,e)]$ falls into the range of $R(\Phi,e)$. Then there exists some $\tilde{e}$ such that $R(\Phi,\tilde{e})=\bbE_{e} [R(\Phi,e)]$. This $\tilde{e}$ can be seen as some mean-risk environment, and $(\tilde{e},\bbE_{e} [R(\Phi,e)])$ is in the midst of the graph $\{(e,R(\Phi,e)): e\in \mE_{tr}\}$. The variance in \eqref{eqn:varRe} actually involves a set of vectors in the graph space $\{(e,R(\Phi,e))-(\tilde{e},\bbE_{e} [R(\Phi,e)]): e\in \mE_{tr}\}$ which can span a linear space to extrapolate risk, as shown in Figure \ref{fig:riskextra}. The performance of such risk extrapolation depends on the diversity of the training environment set $\mE_{tr}$ and the measure $\mu$.

\subsection{Proof of Theorem \ref{prop:ZIN}}
\label{proof:ZINL2}
To prove this theorem, we first absorb \eqref{eqn:ZIN} into the framework of this paper. Suppose the learning model uses a finite set of $E$ fundamental environments (not necessarily known by the model) as the basis defined in Definition \ref{dfn:basisgeneral}: $\mB_{tr}:=\{e_i\}_{i=1}^E$. Then the first term in \eqref{eqn:ZIN} can be rewritten as follows: 
\begin{align}
\label{eqn:ZINrewrite}
 \tilde{R}(w{\circ} \Phi,\frac{1}{E}1_{(E)}){\cdot} 1_{(E)}=\sum_{i=1}^E [\tilde{R}(w{\circ} \Phi,\frac{1}{E}1_{(E)})]_i= \frac{1}{E}\sum_{i=1}^E  R(w{\circ} \Phi,e_i),
\end{align}
where $[\cdot]_i$ denotes the $i$-th dimension of a vector. Here we use $\tilde{R}$ to denote the original form of learning risk in \eqref{eqn:ZIN}, in order to distinguish it from $R$ in our framework. In fact, $\tilde{R}$ is an $E$-dimensional vector that represents the learning risk in the $E$ fundamental environments. This representation is equivalent to interpreting $R$ as a function of $e_i$ with $e_i$ varying in the $E$ fundamental environments. Therefore, \eqref{eqn:ZINrewrite} is true.

Similarly, the second term of \eqref{eqn:ZIN} can be rewritten as follows:
\begin{align}
\label{eqn:ZINrewrite2}
\|\nabla_w \tilde{R}(w{\circ} \Phi,\rho)\|_2^2=\sum_{i=1}^E [|\nabla_w \tilde{R}(w{\circ} \Phi,\rho)|]_i^2=\sum_{i=1}^E|\nabla_w R(w{\circ} \Phi,e_i)|^2 \rho_i,\quad \rho\in \Delta^E.
\end{align}
It further assigns a probability $\rho\in \Delta^E$ (see Eq. \ref{eqn:Esimplex} for the definition) to the fundamental environments, while \eqref{eqn:ZINrewrite} uses the uniform probability $\frac{1}{E}1_{(E)}$. 

Both \eqref{eqn:ZINrewrite} and \eqref{eqn:ZINrewrite2} are essentially mathematical expectations:
\begin{align}
\label{eqn:ZINrewriteexpect}
&\bbE_{\frac{1_{(E)}}{E}} [R(w{\circ} \Phi,e)]:=\int_{\mE_{tr}} R(w{\circ} \Phi,e)\ud(\frac{1_{(E)}}{E})= \frac{1}{E}\sum_{i=1}^E  R(w{\circ} \Phi,e_i), \\
\label{eqn:ZINrewrite2expect}
&\bbE_{\rho} [|\nabla_w R(w{\circ} \Phi,e)|^2]:=\int_{\mE_{tr}}|\nabla_w R(w{\circ} \Phi,e)|^2\drho=\sum_{i=1}^E|\nabla_w R(w{\circ} \Phi,e_i)|^2 \rho_i.
\end{align}
From Theorem \ref{thm:IRMTVL2}, there exist induced probability measures for $w$ such that
\begin{align}
\label{eqn:ZINrewriteexpectw}
\bbE_{w \leftarrow \frac{1_{(E)}}{E}} [ R(w{\circ} \Phi)]=\bbE_{\frac{1_{(E)}}{E}} [R(w{\circ} \Phi,e)],\quad \bbE_{w\leftarrow \rho} [|\nabla_w R(w{\circ} \Phi)|^2 ]=\bbE_{\rho} [|\nabla_w R(w{\circ} \Phi,e)|^2].
\end{align}
Both ZIN and TIVA aim to learn the probability $\rho$ from the training samples with environment information. They differ mainly in their learning approach. Therefore, both ZIN and TIVA can be generalized to \eqref{eqn:MinimaxTVL2}.

\subsection{Proof of Theorem \ref{thm:IRMTVL1}}
\label{proof:IRMTVL1}
The proof is almost identical to that of \ref{proof:thmIRMTVL2} and \ref{proof:ZINL2}, except that $|\nabla_w R(w{\circ} \Phi)|^2 $ is replaced by $|\nabla_w R(w{\circ} \Phi)|$. Note that the condition $|\nabla_w R(w(e)\circ \Phi)|\in L^1(\mE_{tr},\sF_{tr},\mu)$ in Theorem \ref{thm:IRMTVL1} is weaker than $|\nabla_w R(w(e)\circ \Phi)|\in  L^2(\mE_{tr},\sF_{tr},\mu)$ in Theorem \ref{thm:IRMTVL2} when $\mu(\mE_{tr})<\infty$. Therefore, \eqref{eqn:IRMv1wTVL1} and \eqref{eqn:MinimaxTVL1} allow broader classes of $R(w{\circ} \Phi)$ to be used than \eqref{eqn:IRMv1wTVL2} and \eqref{eqn:MinimaxTVL2}.

\subsection{Proof of Proposition \ref{prop:IRMTVL1coarea}}
\label{proof:IRMTVL1coarea}
The key to prove this proposition is to partition the real space into countable continuous intervals such that the level sets of $R(w\circ \Phi)$ w.r.t. these intervals have positive volumes. Specifically, we establish a set of $K\in \bbN^+\cup \{+\infty\}$ intervals
\begin{align}
\label{eqn:RpartK}
&\{B_k:=[a_k,b_k)\}_{k=1}^K  \quad \st \quad a_k<b_k\leqs a_{k+1},\quad  k=1,2,\cdots, K, \nonumber\\
&\qquad R(w\circ \Phi)\ \text{is Lipschitz continuous on}\ \mG_k:=\{w\in \Omega: a_k<R(w\circ \Phi)<b_k\}.
\end{align}
Since $a_k<b_k$, each $B_k$ contains at least one rational number. Besides, since $B_k\cap B_j=\emptyset$ for any $k\ne j$, the rational numbers in $B_k$ cannot overlap those in $B_j$. Since all the rational numbers are dense in $\bbR$ and the volume $|B_k|:=(b_k-a_k)>0$ for all $k$, there are at most countable intervals $\{B_k\}_{k=1}^K$. On the contrary, if none of such interval $B_k$ exists, it is obvious that the integral in \eqref{eqn:IRMTVL1coarea} equals $0$.

Given the conditions, $\int_\Omega |\nabla_w R(w\circ \Phi)| \dnu<\infty$. Since $B_k\subseteq \Omega$ and $|\nabla_w R(w\circ \Phi)|\geqs 0$, $\int_{B_k} |\nabla_w R(w\circ \Phi)| \dnu \leqs \int_\Omega |\nabla_w R(w\circ \Phi)| \dnu<\infty$ holds for all $k$. Thus, we can decompose the integral as follows:
\begin{align}
\label{eqn:intdecomp}
\int_\Omega |\nabla_w R(w\circ \Phi)| \dnu=\int_{\Omega\backslash(\bigcup_{k=1}^K \mG_k)} |\nabla_w R(w\circ \Phi)| \dnu+ \int_{\bigcup_{k=1}^K \mG_k} |\nabla_w R(w\circ \Phi)| \dnu.
\end{align}
Since $\Omega\backslash(\bigcup_{k=1}^K \mG_k)$ corresponds to $\bbR\backslash(\bigcup_{k=1}^K B_k)$ where the function value $R(w\circ \Phi)$ is discontinuous almost everywhere (a.e.), the Lebesgue measure (i.e., the $d$-dimensional volume) $\dnu=0$ on $\Omega\backslash(\bigcup_{k=1}^K \mG_k)$. Hence, the first term in \eqref{eqn:intdecomp} $\int_{\Omega\backslash(\bigcup_{k=1}^K \mG_k)} |\nabla_w R(w\circ \Phi)| \dnu=0$. From \cite{coarealip}, the coarea formula holds for all the Lipschitz continuous parts:
\begin{align}
\label{eqn:coarealip}
\int_{\mG_k} |\nabla_w R(w\circ \Phi)| \dnu=\int_{a_k}^{b_k}\int_{\{w\in \Omega: R(w\circ \Phi)=\gamma\}} \ud s \ud \gamma,\quad  k=1,2,\cdots, K. 
\end{align}
Exploiting the $\sigma$-additivity of Lebesgue measure yields
\begin{align}
\label{eqn:coarealipsum}
\int_{\bigcup_{k=1}^K \mG_k} |\nabla_w R(w\circ \Phi)| \dnu&=\sum_{k=1}^K\int_{\mG_k} |\nabla_w R(w\circ \Phi)| \dnu=\sum_{k=1}^K\int_{a_k}^{b_k}\int_{\{w\in \Omega: R(w\circ \Phi)=\gamma\}} \ud s \ud \gamma\nonumber\\
&=\int_{\bigcup_{k=1}^K B_k}\int_{\{w\in \Omega: R(w\circ \Phi)=\gamma\}} \ud s \ud \gamma. 
\end{align}
Besides, it is also obvious to see that $\ud \gamma=0$ on $\bbR\backslash(\bigcup_{k=1}^K B_k)$ where $\gamma=R(w\circ \Phi)$ is discontinuous a.e. Hence $\int_{\bbR\backslash(\bigcup_{k=1}^K B_k)}\int_{\{w\in \Omega: R(w\circ \Phi)=\gamma\}} \ud s \ud \gamma=0$. Together with \eqref{eqn:intdecomp} and \eqref{eqn:coarealipsum}, we have
\begin{align}
\label{eqn:IRMTVL1coareaproof}
\int_{-\infty}^{\infty}\int_{\{w\in \Omega: R(w\circ \Phi)=\gamma\}} \ud s \ud \gamma&=\int_{\bigcup_{k=1}^K B_k}\int_{\{w\in \Omega: R(w\circ \Phi)=\gamma\}} \ud s \ud \gamma=\int_{\bigcup_{k=1}^K \mG_k} |\nabla_w R(w\circ \Phi)| \dnu\nonumber\\
&=\int_\Omega |\nabla_w R(w\circ \Phi)| \dnu.
\end{align}
It proves the coarea formula in Proposition \ref{prop:IRMTVL1coarea}. 

This proof also explains the mathematical essence of why TV-$\ell_1$ models can learn invariant features. To achieve a smaller TV for $R(w\circ \Phi)$, the continuous interval $B_k$ should take a smaller length, especially on the level set $\{w\in \Omega: R(w\circ \Phi)=\gamma\}$ with a large area. To do this, the optimizing procedure may squash the graph of $R(w\circ \Phi)$ w.r.t. $w$ in every continuous level set $\mG_k$, making the graph more blocky (piece-wise constant). This makes $R(w\circ \Phi)$ more robust w.r.t. $w$ in each $\mG_k$.

\subsection{Proof of Theorem \ref{thm:OODwfrome}}
\label{proof:OODwfrome}
From the implied facts of \eqref{eqn:OOD}, we only need to prove that for any $\Phi$ such that $\max_{e\in \mE_{all}} R(\Phi,e)<\infty$, $\max_{e\in \mE_{all}} R(\Phi,e)=\max_{w\in \Omega_{all}} R(w\circ \Phi)$. From Conditions 2a and 3, it suffices to prove that $\forall \Phi$, $\max_{e\in \mE_{all}} R(w(e)\circ \Phi)=\max_{w\in \Omega_{all}} R(w\circ \Phi)$. First, we examine how the left side yields the right side. Denote $e^*=\argmax_{e\in \mE_{all}} R(w(e)\circ \Phi)$. Then $R(w(e^*)\circ \Phi)\geqs \max_{w\in \Omega_{all}} R(w\circ \Phi)$. If not, then there exists $\omega^\bullet\in \Omega_{all}$ such that $R(w(e^*)\circ \Phi)<R(\omega^\bullet\circ \Phi)$. Since $w$ is a surjective mapping, we have $w^{-1}(\omega^\bullet)\ne \emptyset$ and $e^*\notin w^{-1}(\omega^\bullet)$. Picking up any $e^\bullet\in w^{-1}(\omega^\bullet)$, we have $R(w(e^\bullet)\circ \Phi)>R(w(e^*)\circ \Phi)$, which violates the optimality of $e^*$.

Second, we examine how the right side yields the left side. Denote $w^*=\argmax_{w\in \Omega_{all}} R(w\circ \Phi)$. Then $R(w^*\circ \Phi)\geqs \max_{e\in \mE_{all}} R(w(e)\circ \Phi)$. If not, then there exists $e^\bullet\in \mE_{all}$ such that $R(w^*\circ \Phi)< R(w(e^\bullet)\circ \Phi)$. Hence $R(w^*\circ \Phi)< R(w^\bullet\circ \Phi)$ with $w^\bullet:=w(e^\bullet)\in \Omega_{all}$, which violates the optimality of $w^*$.

Summarizing the inequalities of both directions, we have $\max_{e\in \mE_{all}} R(w(e)\circ \Phi)=\max_{w\in \Omega_{all}} R(w\circ \Phi)$, $\forall \Phi$. Thus minimizing either side of this equality w.r.t. $\Phi$ leads to the same optimization model.

\subsection{Proof of Theorem \ref{thm:OODwequiv}}
\label{proof:OODwequiv}

\subsubsection{Allowing $\lambda$ to Vary with $\Phi$}
\label{proof:OODwequivvar}
To prove that $\lambda$ should be allowed to vary with $\Phi$, we only need to provide a counterexample with a fixed $\lambda$. Consider a simple function fitting problem that uses $w\cdot \Phi$ to fit the constant function $-1$. We use the absolute fitting error as the learning risk $R(w\circ \Phi):=|w\cdot \Phi-(-1)|=|w\cdot \Phi+1|$. Suppose we aim to learn the best feature parameter $\Phi\in [-1,1]$. The classifier $w$ lies uniformly in $[-0.9,0.1]$, but its center deviates from $0$ after conveying the environment information $e$. Equip $w$ with the uniform probability measure on $[-0.9,0.1]$ as $\nu$. Now $R(w\circ \Phi)=1+w\cdot \Phi$ according to the domains of $w$ and $\Phi$. It can be easily calculated that
\begin{align}
\label{eqn:Rwfixed}
&\max_{w\in \Omega_{all}} R(w{\circ} \Phi)=\max_{w\in [-0.9,0.1]} \{1+w\cdot \Phi\}=\left\{\begin{array}{ll}
1+0.1\Phi & \text{if}\quad \Phi\geqs 0\\
1-0.9\Phi & \text{if}\quad \Phi<0
\end{array} \right.,\\
\label{eqn:Rwfixed2}
&\min_{\Phi\in [-1,1]}\max_{w\in \Omega_{all}} R(w{\circ} \Phi)= 1, \quad \argmin_{\Phi\in [-1,1]}\max_{w\in \Omega_{all}} R(w{\circ} \Phi)=0. 
\end{align}
However,
\begin{align}
\label{eqn:IRMTVL1fixed}
&\bbE_{w}^{all} [R(w\circ \Phi)] +\lambda  (\bbE_{w}^{all} [|\nabla_w R(w\circ \Phi)| ])^2=\int_{-0.9}^{0.1} (1+w\cdot \Phi) \dnu+\lambda(|\Phi|\int_{-0.9}^{0.1}  \dnu)^2=1-\frac{2}{5}\Phi+\lambda \Phi^2,\\
\label{eqn:IRMTVL1fixed2}
&\min_{\Phi\in [-1,1]}\{\bbE_{w}^{all} [R(w\circ \Phi)] +\lambda  (\bbE_{w}^{all} [|\nabla_w R(w\circ \Phi)| ])^2\}=\left\{\begin{array}{ll}
1-\frac{1}{25\lambda} & \text{if}\quad \lambda>\frac{1}{5}\\
\frac{3}{5}+\lambda & \text{if}\quad 0<\lambda\leqs\frac{1}{5}\\
\frac{3}{5} & \text{if}\quad \lambda= 0
\end{array} \right.,    \nonumber\\
&\argmin_{\Phi\in [-1,1]}\{\bbE_{w}^{all} [R(w\circ \Phi)] +\lambda  (\bbE_{w}^{all} [|\nabla_w R(w\circ \Phi)| ])^2\}=\left\{\begin{array}{ll}
\frac{1}{5\lambda} & \text{if}\quad \lambda>\frac{1}{5}\\
1 & \text{if}\quad 0\leqs\lambda\leqs\frac{1}{5}
\end{array} \right..
\end{align}
Comparing \eqref{eqn:IRMTVL1fixed2} with \eqref{eqn:Rwfixed2}, we have
\begin{align}
\label{eqn:RwfixedneIRMTVL1fixed}
&\min_{\Phi\in [-1,1]}\{\bbE_{w}^{all} [R(w\circ \Phi)] +\lambda  (\bbE_{w}^{all} [|\nabla_w R(w\circ \Phi)| ])^2\}\ne\min_{\Phi\in [-1,1]}\max_{w\in \Omega_{all}} R(w{\circ} \Phi),\nonumber\\
&\argmin_{\Phi\in [-1,1]}\{\bbE_{w}^{all} [R(w\circ \Phi)] +\lambda  (\bbE_{w}^{all} [|\nabla_w R(w\circ \Phi)| ])^2\}\ne \argmin_{\Phi\in [-1,1]}\max_{w\in \Omega_{all}} R(w{\circ} \Phi),
\end{align}
for any fixed $\lambda\geqs 0$. Hence \eqref{eqn:IRMv1wTVL1glob} cannot achieve \eqref{eqn:OODwfrome} in terms of the objective value and the argument.

This counterexample can be explained as follows. The learning risk $R$ increases with $|\Phi|$ for the worst-case $w$. Therefore, $\Phi=0$ is the optimal point in \eqref{eqn:OODwfrome}. However, the deviation of $w$ from $0$ also deviates the expectation of $R$ w.r.t. $w$, leading to the deviation of optimal $\Phi$ from $0$ in \eqref{eqn:IRMv1wTVL1glob}. This deviation of $\Phi$ cannot be offset by the TV term in \eqref{eqn:IRMv1wTVL1glob}, no matter how we set the fixed $\lambda$.

\subsubsection{Existence of $\lambda_\Phi$}
First, we consider the trivial case $\bbE_{w}^{all} [|\nabla_w R(w\circ \Phi)| ]=0$, which indicates a zero TV term. When $R(w\circ \Phi)$ is Lipschitz continuous w.r.t. $w$, $\bbE_{w}^{all} [|\nabla_w R(w\circ \Phi)| ]=0$ implies that $R(w\circ \Phi)$ is constant w.r.t. $w$. In this case, $\bbE_{w}^{all} [R(w\circ \Phi)]=\max_{w\in \Omega_{all}} R(w{\circ} \Phi)$ and $\lambda$ can be set arbitrarily. 

Next, we investigate the nontrivial case $\bbE_{w}^{all} [|\nabla_w R(w\circ \Phi)| ]>0$. We have
\begin{align}
\label{eqn:Rwsmallermax}
\bbE_{w}^{all} [R(w\circ \Phi)]\leqs \bbE_{w}^{all} [\max_{\omega\in \Omega_{all}} R(\omega{\circ} \Phi)]\leqs(\max_{\omega\in \Omega_{all}} R(\omega{\circ} \Phi))\cdot\bbE_{w}^{all}[1]=\max_{\omega\in \Omega_{all}} R(\omega{\circ} \Phi).
\end{align}
Note that $\max_{\omega\in \Omega_{all}} R(\omega{\circ} \Phi)$ is constant w.r.t. $w$ and thus can be moved outside the expectation. Then $\lambda_\Phi$ exists by the form:
\begin{align}
\label{eqn:lbdphidef}
\lambda_\Phi:=\frac{(\max_{w\in \Omega_{all}} R(w{\circ} \Phi))-\bbE_{w}^{all} [R(w\circ \Phi)]}{(\bbE_{w}^{all} [|\nabla_w R(w\circ \Phi)| ])^2}\geqs 0.
\end{align}
This quotient depends only on $\Phi$, but not $w$.

\subsubsection{Achieving Optimality in \eqref{eqn:OODwfrome}}
$\Phi^\bullet$ being the optimal point of \eqref{eqn:IRMv1wTVL1globphi} indicates
\begin{align}
\label{eqn:solIRMv1wTVL1globphi}
\bbE_{w}^{all} [R(w\circ \Phi^\bullet)] +\lambda_\Phi^\bullet  (\bbE_{w}^{all} [|\nabla_w R(w\circ \Phi^\bullet)| ])^2\leqs \bbE_{w}^{all} [R(w\circ \Phi)] +\lambda_\Phi  (\bbE_{w}^{all} [|\nabla_w R(w\circ \Phi)| ])^2,\ \forall \Phi.
\end{align}
In particular, $\Phi$ can be any optimal point $\Phi^*$ of \eqref{eqn:OODwfrome}. Together with \eqref{eqn:OODwequal}, we have
\begin{align}
\label{eqn:solIRMv1wTVL1globphioptim}
\max_{w\in \Omega_{all}} R(w{\circ} \Phi^\bullet)&=\bbE_{w}^{all} [R(w\circ \Phi^\bullet)] +\lambda_\Phi^\bullet  (\bbE_{w}^{all} [|\nabla_w R(w\circ \Phi^\bullet)| ])^2\nonumber\\
&\leqs \bbE_{w}^{all} [R(w\circ \Phi^*)] +\lambda_\Phi^*  (\bbE_{w}^{all} [|\nabla_w R(w\circ \Phi^*)| ])^2=\max_{w\in \Omega_{all}} R(w{\circ} \Phi^*).
\end{align}
Hence $\Phi^\bullet$ is also an optimal point of \eqref{eqn:OODwfrome}. 

Conversely, $\Phi^*$ being the optimal point of \eqref{eqn:OODwfrome} indicates
\begin{align}
\label{eqn:solIRMv1wTVL1globphioptiminv}
\bbE_{w}^{all} [R(w\circ \Phi^*)] &+\lambda_\Phi^*  (\bbE_{w}^{all} [|\nabla_w R(w\circ \Phi^*)| ])^2=\max_{w\in \Omega_{all}} R(w{\circ} \Phi^*)\nonumber\\
&\leqs \max_{w\in \Omega_{all}} R(w{\circ} \Phi^\bullet)=\bbE_{w}^{all} [R(w\circ \Phi^\bullet)] +\lambda_\Phi^\bullet  (\bbE_{w}^{all} [|\nabla_w R(w\circ \Phi^\bullet)| ])^2.
\end{align}
Hence $\Phi^*$ is also an optimal point of \eqref{eqn:IRMv1wTVL1globphi}.

\subsection{Proof of Theorem \ref{thm:OODwequivmm}}
\label{proof:OODwequivmm}

\subsubsection{Allowing $\lambda$ to Vary with $\Phi$}
The counterexample is similar to that in \ref{proof:OODwequivvar} except for some changes. First, the measure $\nu_{\frac{1_{(E)}}{E}}$ for $w$ induced by $\frac{1_{(E)}}{E}$ can be set as the uniform probability measure on $[-0.9,0.1]$. The measure $\nu_{\rho}$ for $w$ induced by $\rho$ can be any arbitrary probability measure. Then 
\begin{align}
\label{eqn:MinimaxTVL1fixed}
&\bbE_{w \leftarrow \frac{1_{(E)}}{E}}^{all} [ R(w{\circ} \Phi)] {+}\lambda  \max_{\rho}(\bbE_{w\leftarrow \rho}^{all} [|\nabla_w R(w{\circ} \Phi)| ])^2\nonumber\\
=&\int_{-0.9}^{0.1} (1+w\cdot \Phi) \dnu_{\frac{1_{(E)}}{E}}+\lambda \max_{\rho} (|\Phi|\int_{-0.9}^{0.1}  \dnu_{\rho})^2=1-\frac{2}{5}\Phi+\lambda \Phi^2,
\end{align}
which is the same as that in \eqref{eqn:IRMTVL1fixed}. Similar results to \eqref{eqn:IRMTVL1fixed2} and \eqref{eqn:RwfixedneIRMTVL1fixed} also hold for any fixed $\lambda\geqs 0$. Thus \eqref{eqn:OODwfrome} cannot be achieved by \eqref{eqn:MinimaxTVL1glob}.

\subsubsection{Existence of $\lambda_\Phi$}
If $\bbE_{w\leftarrow \rho}^{all} [|\nabla_w R(w{\circ} \Phi)| ]= 0$ for all $\rho$ and $R(w{\circ} \Phi)$ is Lipschitz continuous w.r.t. $w$, then $R(w{\circ} \Phi)$ is constant w.r.t. $w$. In this case, $\bbE_{w \leftarrow \frac{1_{(E)}}{E}}^{all} [ R(w{\circ} \Phi)]=\max_{w\in \Omega_{all}} R(w{\circ} \Phi)$ and $\lambda$ can be set arbitrarily. 

If $\bbE_{w\leftarrow \rho}^{all} [|\nabla_w R(w{\circ} \Phi)| ]> 0$ for some $\rho$, then $\max_{\rho}(\bbE_{w\leftarrow \rho}^{all} [|\nabla_w R(w{\circ} \Phi)| ])^2>0$. Similar to \eqref{eqn:Rwsmallermax}, 
\begin{align}
\label{eqn:Rrhosmallermax}
\bbE_{w \leftarrow \frac{1_{(E)}}{E}}^{all} [ R(w{\circ} \Phi)]\leqs \max_{\omega\in \Omega_{all}} R(\omega{\circ} \Phi).
\end{align}
Then $\lambda_\Phi$ exists by the form:
\begin{align}
\label{eqn:lbdphidefmm}
\lambda_\Phi:=\frac{(\max_{w\in \Omega_{all}} R(w{\circ} \Phi))-\bbE_{w \leftarrow \frac{1_{(E)}}{E}}^{all} [ R(w{\circ} \Phi)]}{\max_{\rho}(\bbE_{w\leftarrow \rho}^{all} [|\nabla_w R(w{\circ} \Phi)| ])^2}\geqs 0.
\end{align}
This quotient depends only on $\Phi$, but not $w$.

\subsubsection{Achieving Optimality in \eqref{eqn:OODwfrome}}
An optimal point $\Phi^\bullet$ of \eqref{eqn:MinimaxTVL1globphi} satisfies
\begin{align}
\label{eqn:solMinimaxTVL1globphi}
\bbE_{w \leftarrow \frac{1_{(E)}}{E}}^{all} [ R(w{\circ} \Phi^\bullet)] {+}\lambda_\Phi^\bullet  \max_{\rho}(\bbE_{w\leftarrow \rho}^{all} [|\nabla_w R(w{\circ} \Phi^\bullet)| ])^2 \leqs \bbE_{w \leftarrow \frac{1_{(E)}}{E}}^{all} [ R(w{\circ} \Phi)] {+}\lambda_\Phi  \max_{\rho}(\bbE_{w\leftarrow \rho}^{all} [|\nabla_w R(w{\circ} \Phi)| ])^2,\ \forall \Phi.
\end{align}
$\Phi$ can be any optimal point $\Phi^*$ of \eqref{eqn:OODwfrome}. By combining \eqref{eqn:OODwequalmm}, we can deduce that
\begin{align}
\label{eqn:solMinimaxTVL1globphioptim}
\max_{w\in \Omega_{all}} R(w{\circ} \Phi^\bullet)&=\bbE_{w \leftarrow \frac{1_{(E)}}{E}}^{all} [ R(w{\circ} \Phi^\bullet)] {+}\lambda_\Phi^\bullet  \max_{\rho}(\bbE_{w\leftarrow \rho}^{all} [|\nabla_w R(w{\circ} \Phi^\bullet)| ])^2\nonumber\\
&\leqs\bbE_{w \leftarrow \frac{1_{(E)}}{E}}^{all} [ R(w{\circ} \Phi^*)] {+}\lambda_\Phi^*  \max_{\rho}(\bbE_{w\leftarrow \rho}^{all} [|\nabla_w R(w{\circ} \Phi^*)| ])^2=\max_{w\in \Omega_{all}} R(w{\circ} \Phi^*).
\end{align}
Hence $\Phi^\bullet$ is also an optimal point of \eqref{eqn:OODwfrome}. 

Conversely, an optimal point $\Phi^*$ of \eqref{eqn:OODwfrome} satisfies
\begin{align}
\label{eqn:solMinimaxTVL1globphioptiminv}
\bbE_{w \leftarrow \frac{1_{(E)}}{E}}^{all} [ R(w{\circ} \Phi^*)] &+\lambda_\Phi^*  \max_{\rho}(\bbE_{w\leftarrow \rho}^{all} [|\nabla_w R(w{\circ} \Phi^*)| ])^2=\max_{w\in \Omega_{all}} R(w{\circ} \Phi^*)\nonumber\\
&\leqs \max_{w\in \Omega_{all}} R(w{\circ} \Phi^\bullet)=\bbE_{w \leftarrow \frac{1_{(E)}}{E}}^{all} [ R(w{\circ} \Phi^\bullet)] {+}\lambda_\Phi^\bullet  \max_{\rho}(\bbE_{w\leftarrow \rho}^{all} [|\nabla_w R(w{\circ} \Phi^\bullet)| ])^2.
\end{align}
Hence $\Phi^*$ is also an optimal point of \eqref{eqn:MinimaxTVL1globphi}.

\subsection{Proof of Theorem \ref{thm:OODminimum}}
\label{proof:OODminimum}
We denote the probability space for the training environment set as $(\mE_{tr},\sF_{tr},\mu_{tr})$ and its induced probability space for $w$ as $(\Omega_{tr},\sF_{\Omega_{tr}},\nu_{tr})$. Since $\mE_{all}\in\sF_{tr}$, we have $\mE_{all}\subseteq \mE_{tr}$. Define $\sF_{tr|all}$ as the restricted $\sigma$-algebra on $\mE_{all}$:
\begin{align}
\label{eqn:restrictalgebra}
\sF_{tr|all}:=\{A\cap \mE_{all}: A\in \sF_{tr}\}.
\end{align}
Since $\mE_{all}\in\sF_{tr}$, we have $A\cap \mE_{all} \in \sF_{tr}$. Hence $\sF_{tr|all}$ is a sub-$\sigma$-algebra of $\sF_{tr}$, and it can still use the probability measure $\mu_{tr}$. Then $(\mE_{all},\sF_{tr|all},\mu_{tr})$ can be used as a coarse probability space for $\mE_{all}$, and its induced probability space for $w$ is $(\Omega_{all},\sF_{\Omega_{tr|all}},\nu_{tr})$, where $\Omega_{all}\subseteq\Omega_{tr}$ and $\sF_{\Omega_{tr|all}}\subseteq \sF_{\Omega_{tr}}$.

Since the learning risk $R(w{\circ} \Phi)$ and $|\nabla_w R(w\circ \Phi)|$ are defined on $(\Omega_{tr},\sF_{\Omega_{tr}},\nu_{tr})$, the following objective function and optimization model are well-defined:
\begin{align}
\label{eqn:allbytrobj}
& \int_{\Omega_{all}} R(w{\circ} \Phi)\dnu_{tr} {+}\lambda_\Phi  (\int_{\Omega_{all}} |\nabla_w R(w{\circ} \Phi)|\dnu_{tr} )^2=: \bbE_{w}^{tr} [R(w\circ \Phi)] +\lambda_\Phi  (\bbE_{w}^{tr} [|\nabla_w R(w\circ \Phi)| ])^2,\\
\label{eqn:IRMv1wTVL1trainphi}
&\min_{\Phi}\ \left\{ \bbE_{w}^{tr} [R(w\circ \Phi)] +\lambda_\Phi  (\bbE_{w}^{tr} [|\nabla_w R(w\circ \Phi)| ])^2 \right\}. \tag{IRM-TV-$\ell_1$-training-$\Phi$}
\end{align}
The only difference of \eqref{eqn:IRMv1wTVL1trainphi} from \eqref{eqn:IRMv1wTVL1} is the variable $\lambda_\Phi$, which does not depend on $w$ and its probability space. It can be easily verified that \eqref{eqn:IRMv1wTVL1trainphi} has the same properties and capabilities as \eqref{eqn:IRMv1wTVL1globphi} in Theorem \ref{thm:OODwequiv}.

As for \eqref{eqn:MinimaxTVL1}, we denote the probability space for $w$ induced by $\rho$ as $(\Omega_{tr},\sF_{\Omega_{tr}},\nu_{tr}^\rho)$. Following the above procedure, we obtain $(\Omega_{all},\sF_{\Omega_{tr|all}},\nu_{tr}^\rho)$ as a coarse probability space for all $\rho$. Then the following objective function and optimization model are also well-defined:
\begin{align}
\label{eqn:allbytrMinimaxobj}
& \int_{\Omega_{all}} R(w{\circ} \Phi)\dnu_{tr}^{\frac{1_{(E)}}{E}} {+}\lambda_\Phi \max_{\rho} (\int_{\Omega_{all}} |\nabla_w R(w{\circ} \Phi)|\dnu_{tr}^\rho )^2=: \bbE_{w \leftarrow \frac{1_{(E)}}{E}}^{tr} [ R(w{\circ} \Phi)] {+}\lambda_\Phi  \max_{\rho}(\bbE_{w\leftarrow \rho}^{tr} [|\nabla_w R(w{\circ} \Phi)| ])^2,\\
\label{eqn:MinimaxTVL1trainphi}
&\min_{\Phi}\ \left\{ \bbE_{w \leftarrow \frac{1_{(E)}}{E}}^{tr} [ R(w{\circ} \Phi)] {+}\lambda_\Phi  \max_{\rho}(\bbE_{w\leftarrow \rho}^{tr} [|\nabla_w R(w{\circ} \Phi)| ])^2 \right\}. \tag{Minimax-TV-$\ell_1$-training-$\Phi$}
\end{align}
Similarly, \eqref{eqn:MinimaxTVL1trainphi} has the same properties and capabilities as \eqref{eqn:MinimaxTVL1globphi} in Theorem \ref{thm:OODwequivmm}.

Although \eqref{eqn:IRMv1wTVL1trainphi} and \eqref{eqn:MinimaxTVL1trainphi} have the same capabilities as \eqref{eqn:IRMv1wTVL1globphi} and \eqref{eqn:MinimaxTVL1globphi} to achieve OOD generalization, the former two are not necessarily equivalent to the latter two. Denote the probability space for the global environment set as $(\mE_{all},\sF_{all},\mu_{all})$ and its induced probability space for $w$ as $(\Omega_{all},\sF_{\Omega_{all}},\nu_{all})$. Let $\sF_{\Omega_{tr\cap all}}:=\sF_{\Omega_{tr|all}}\cap \sF_{\Omega_{all}}$ be the intersection of the two $\sigma$-algebras, which is also a $\sigma$-algebra. It can be anticipated that $\nu_{tr}(A)=\nu_{all}(A)$, $\forall A\in \sF_{\Omega_{tr\cap all}}$. That is, the measure for both training and test environments should be the same. However, calculating the integrals of $R(w{\circ} \Phi)$ and $|\nabla_w R(w\circ \Phi)|$ on $\sF_{\Omega_{tr\cap all}}$ may result in distortion when $\sF_{\Omega_{tr\cap all}}\subsetneqq \sF_{\Omega_{tr|all}}$. Because for some $A\in \sF_{\bbR}$, $\{w: R(w{\circ} \Phi)\in A\}\in \sF_{\Omega_{tr|all}}\backslash \sF_{\Omega_{tr\cap all}}$ and thus $\bbI_{\{w: R(w{\circ} \Phi)\in A\}}$ cannot be used in \ref{proof:thmIRMTVL2integral} to construct $\int_{\Omega_{all}} R(w{\circ} \Phi)\dnu_{tr}$.

Another similar case is $\Omega_{tr}=\Omega_{all}$ but $\sF_{\Omega_{tr}}\subsetneqq\sF_{\Omega_{all}}$. In this scenario, calculating the integrals of $R(w{\circ} \Phi)$ and $|\nabla_w R(w\circ \Phi)|$ on $\sF_{\Omega_{tr}}$ may result in distortion of \eqref{eqn:IRMv1wTVL1globphi} and \eqref{eqn:MinimaxTVL1globphi}. This is because for some $A\in \sF_{\bbR}$, $\{w: R(w{\circ} \Phi)\in A\}\in \sF_{\Omega_{all}}\backslash \sF_{\Omega_{tr}}$ and thus $\bbI_{\{w: R(w{\circ} \Phi)\in A\}}$ cannot be used in \ref{proof:thmIRMTVL2integral} to construct $\int_{\Omega_{all}} R(w{\circ} \Phi)\dnu_{all}$. 

Last, if $\sF_{\Omega_{tr}}\supseteq\sF_{\Omega_{all}}$, then \eqref{eqn:IRMv1wTVL1trainphi} and \eqref{eqn:MinimaxTVL1trainphi} imply \eqref{eqn:IRMv1wTVL1globphi} and \eqref{eqn:MinimaxTVL1globphi}, as concluded in Corollary \ref{cor:OODsufficient}.

\setcounter{table}{0}
\renewcommand{\thetable}{B\arabic{table}}

\setcounter{figure}{0}
\renewcommand{\thefigure}{B\arabic{figure}}

\section{Implementation Details}\label{implementation}

\subsection{Subgradient Approach to Solve IRM-TV-$\ell_1$ and Minimax-TV-$\ell_1$}
\label{sec:IRMTVL1solve}
First, we give the definition of the Fr{\'e}chet subdifferential of $f: \bbRdmY\rightarrow \bbR$ at $w$ denoted by $\partial f(w)$:
\begin{definition}[The Fr{\'e}chet Subdifferential]
\label{def:frechetsubdifferential}
\begin{align}
\label{eqn:frechetsubdifferential}
\partial f(w){:=} \left\{v{\in}\bbRdmY: \liminf_{\substack{u\to w\\ u\neq w}}\frac{f(u){-}f(w){-}v\cdot(u{-}w)}{\|u-w\|_2}{\geqs}0\right\}.
\end{align}
\end{definition}
The subgradient of the modulus function (i.e., $\ell_2$ norm) can be easily calculated:
\begin{align}
\label{eqn:subgradmod}
\partial |w|{=} \left\{\begin{array}{ll}
\frac{w}{|w|} & \text{if}\quad w\ne 0,\\
\{v{\in}\bbRdmY:  |v|\leqs 1\} & \text{if}\quad  w=0.
\end{array} \right.
\end{align}
In this paper, we directly take $0$ as the convenient subgradient for $\partial |w|$ at $w=0$, then $\partial |w|$ can be used in a similar way to a gradient. In particular, the chain rule yields:
\begin{align}
\label{eqn:subgradnabR}
&\partial_{\Phi} |\nabla_w R(w\circ \Phi)|{=}\left\{\begin{array}{ll}
\frac{J_{\Phi}^\top[\nabla_w R(w\circ \Phi)]*\nabla_w R(w\circ \Phi)}{|\nabla_w R(w\circ \Phi)|} & \text{if}\quad \nabla_w R(w\circ \Phi)\ne 0,\\
0  &  \text{if}\quad   \nabla_w R(w\circ \Phi)=0,
\end{array} \right.
\end{align}
where $J_{\Phi}^\top[\cdot]$ denotes the transpose of the Jacobian matrix w.r.t. $\Phi$, and $*$ denotes the matrix multiplication. For a simpler form with a one-dimensional $w$,
\begin{align}
\label{eqn:subgradnabR1dim}
&\partial_{\Phi} | R'_w(w\circ \Phi)|{=} \sign (R'_w(w\circ \Phi)) \nabla_{\Phi}[R'_w(w\circ \Phi)].
\end{align}

The objective function in \eqref{eqn:IRMv1wTVL1} and its subgradient are
\begin{align}
\label{eqn:IRMTVL1obj}
g(\Phi)&=\bbE_{w} [R(w\circ \Phi)] +\lambda  (\bbE_{w} [|\nabla_w R(w\circ \Phi)| ])^2,\qquad\\
\label{eqn:IRMTVL1objsubgrad}
\partial_{\Phi} g(\Phi) &=\bbE_{w} [\nabla_{\Phi} R(w\circ \Phi)] +2\lambda\bbE_{w} [|\nabla_w R(w\circ \Phi)| ]\bbE_{w} [\partial_{\Phi}|\nabla_w R(w\circ \Phi)| ].
\end{align}
Then the update formula for $\Phi$ at the $(k+1)$-th iteration is given by the following equation, where $\eta>0$ is the learning rate:
\begin{align}
\label{eqn:IRMTVL1update}
\Phi^{(k+1)}=\Phi^{(k)}-\eta \partial_{\Phi} g(\Phi^{(k)}).
\end{align}

We adopt the adversarial training architecture of \cite{ZIN} to solve \eqref{eqn:MinimaxTVL1}. Its objective function, gradient and subgradient are
\begin{align}
\label{eqn:MinimaxTVL1obj}
h(\Phi,\rho) =& \bbE_{w \leftarrow \frac{1_{(E)}}{E}} [ R(w{\circ} \Phi)] {+}\lambda (\bbE_{w\leftarrow \rho} [|\nabla_w R(w{\circ} \Phi)| ])^2,\\
\label{eqn:MinimaxTVL1objgrad}
\nabla_{\rho} h(\Phi,\rho)=& 2\lambda \bbE_{w{\leftarrow} \rho} [|\nabla_w R(w{\circ} \Phi)| ] \nabla_{\rho}(\bbE_{w{\leftarrow} \rho} [|\nabla_w R(w{\circ} \Phi)| ]),\\
\label{eqn:MinimaxTVL1objsubgrad}
\partial_{\Phi} h(\Phi,\rho)=& \bbE_{w \leftarrow \frac{1_{(E)}}{E}} [ \nabla_{\Phi}R(w{\circ} \Phi)]{+}2\lambda\bbE_{w{\leftarrow} \rho} [|\nabla_w R(w\circ \Phi)| ]\bbE_{w{\leftarrow} \rho} [\partial_{\Phi}|\nabla_w R(w\circ \Phi)| ].
\end{align}
Then the update formulae for $\rho$ and $\Phi$ at the $(k+1)$-th iteration are given by the following equations, where $\eta_1,\eta_2>0$ are the learning rates:
\begin{align}
\label{eqn:MinimaxTVL1update2}
\rho^{(k+1)}&=\rho^{(k)}+\eta_1 \nabla_{\rho} h(\Phi^{(k)},\rho^{(k)}),\\
\label{eqn:MinimaxTVL1update1}
\Phi^{(k+1)}&=\Phi^{(k)}-\eta_2 \partial_{\Phi} h(\Phi^{(k)},\rho^{(k+1)}).
\end{align}

We use the Adam scheme \cite{adam} for Pytorch\footnote{https://pytorch.org} as the optimizer. We apply min-batch subgradients with batch size 1024 in Landcover, and full-batch subgradients in the other data sets. More implementing details can be found in the code link, such as the learning rate, the number of training epochs, etc.

To verify that \ref{eqn:IRMv1wTVL1} (or \ref{eqn:MinimaxTVL1}) will not induce higher computational complexity than \ref{eqn:IRMv1wTVL2} (or \ref{eqn:ZIN}/\ref{eqn:MinimaxTVL2}), we run \ref{eqn:IRMv1wTVL1}, \ref{eqn:MinimaxTVL1}, \ref{eqn:IRMv1wTVL2} and \ref{eqn:ZIN} for $5300$ epochs and $10$ repetitions, then report the average running times (in seconds) in the following Table \ref{tb:complexity}. The STDs are all less than $1.5$ seconds. It indicates that \ref{eqn:IRMv1wTVL1} (or \ref{eqn:MinimaxTVL1}) has almost the same average running times as those of \ref{eqn:IRMv1wTVL2} (or \ref{eqn:ZIN}).

\begin{table}[h]
\caption{Average running times (in seconds) of IRM-TV models by 10 repetitions. }
\label{tb:complexity}
%\vskip 0.15in
\begin{center}
\begin{small}
\begin{sc}
\begin{tabular}{cccccc}
\toprule
 Method & Simulation & House Prices & CelebA &  Landcover  &  Adult \\
\hline
%\abovespace
 ZIN 				& 13.5424  &    18.7798   &  23.5480 &   68.0905  &  32.6906  \\
 Minimax-TV-$\ell_1$   	& 12.8399  &  18.7685     &  23.2884 &   68.3897  &  33.1776  \\
 IRM-TV-$\ell_2$ 		& 8.9668   &   8.9001     &  22.9603 &    N/A  & 11.1240   \\
 IRM-TV-$\ell_1$ 		& 8.4198  &   9.4552     &  24.0967 &    N/A  &  11.7834  \\
\bottomrule
\end{tabular}
\end{sc}
\end{small}
\end{center}
%\vskip -0.1in
\end{table}

\subsection{Synthetic Data Set Generation}
\label{imp:gensyn}
We follow \cite{ZIN,TIVA} to generate synthetic data sets. Let $t$ be the time in $[0, 1]$, and $X_{v}(t)\in\mathbb{R}$ and $X_{s}(t)\in\mathbb{R}$ be the invariant and spurious features, respectively. We have
\begin{equation*}
X_{v}(t)\sim\begin{cases}
\mathcal{N}(1, 1), ~\text{w.p.}~0.5; \\ \mathcal{N}(-1, 1), \text{w.p.}~0.5. 
\end{cases} 
Y(t)\sim\begin{cases}
\text{sign}(X_{v}(t)), &\text{w.p.}~p_{v}; \\ -\text{sign}(X_{v}(t)), &\text{w.p.}~1-p_{v}.
\end{cases} 
X_{s}(t)\sim\begin{cases}
\mathcal{N}(Y(t), 1), &\text{w.p.}~p_{s}(t); \\ \mathcal{N}(-Y(t), 1), &\text{w.p.}~1-p_{s}(t).
\end{cases}
\end{equation*}
Moreover, $X_{v}(t)$ and $X_{s}(t)$ are extended to $5$ and $10$ dimensional sequences by adding up a standard Gaussian noise, respectively.

\subsection{Architectures of $\Phi$ and $\rho$}
We follow \cite{ZIN,TIVA} to instantiate the Pytorch-style architectures of the invariant feature extractor $\Phi$ and the environment inferring measure $\rho$ on each data set, shown in Table \ref{architecture}. $\Phi$ takes primary variables to learn invariant features, while $\rho$ takes auxiliary variables to learn environment partition. All the compared methods except HRM and TIVA share the same $\Phi$, while HRM and TIVA adopt a linear model for $\Phi$. \ref{eqn:MinimaxTVL1} and \ref{eqn:ZIN} share the same $\rho$, while TIVA directly uses auxiliary variables as primary variables in $\Phi$. The cross entropy and the mean squared error are used as prediction loss functions for classification tasks and regression tasks, respectively.

\begin{table}[h]
\caption{Pytorch-style architectures of the invariant feature extractor $\Phi$ and the environment inferring measure $\rho$.}
\label{architecture}
\vskip 0.15in
\begin{center}
\begin{small}
\begin{sc}
\scalebox{0.96}{\begin{tabular}{@{}ccc@{}}
\hline
\abovespace\belowspace
Data Set & $\Phi$ & $\rho$ \\
\hline
\abovespace\belowspace
Simulation  & {Linear(15, 1)} & {Linear(1, 16)$\rightarrow$ReLu()$\rightarrow$Linear(16, 1) $\rightarrow$Sigmoid() }\\
\abovespace\belowspace
House Prices  & Linear(15, 32)$\rightarrow$ReLu()$\rightarrow$Linear(32, 1) & Linear(1, 64)$\rightarrow$ReLu()$\rightarrow$Linear(64, 4)$\rightarrow$Softmax() \\ 
\abovespace\belowspace
CelebA & Linear(512, 16)$\rightarrow$ReLu()$\rightarrow$Linear(16, 1) & Linear(7, 16)$\rightarrow$ReLu()$\rightarrow$Linear(16, 1)$\rightarrow$Sigmoid() \\ 
\abovespace
\multirow{4}{*}{Landcover} & Conv1d(8, 32, 5)$\rightarrow$ReLu()$\rightarrow$Conv1d(32, 32, 3) & \multirow{4}{*}{Linear(2, 16)$\rightarrow$ReLu()$\rightarrow$Linear(16, 2)$\rightarrow$Softmax()}\\
   & $\rightarrow$ReLu()$\rightarrow$MaxPool1d(2, 2)$\rightarrow$Conv1d(32, 64, 3)    &    \\
     & $\rightarrow$ReLu()$\rightarrow$MaxPool1d(2, 2)$\rightarrow$Conv1d(64, 6, 3)  &   \\
\belowspace
     & $\rightarrow$ReLu()$\rightarrow$AvePool1d(1)  &   \\
\abovespace\belowspace
Adult  & Linear(59, 16)$\rightarrow$ReLu()$\rightarrow$Linear(16, 1) & Linear(6, 16)$\rightarrow$ReLu()$\rightarrow$Linear(16, 4)$\rightarrow$Softmax() \\ 
\hline
\end{tabular}}
\end{sc}
\end{small}
\end{center}
\vskip -0.1in
\end{table}

\section{Invariant Feature Identifiability}
\label{sec:identifiable}
\setcounter{table}{0}
\renewcommand{\thetable}{C\arabic{table}}
\setcounter{figure}{0}
\renewcommand{\thefigure}{C\arabic{figure}}

\begin{figure*}[h]
\centering
\subfloat[IRM-TV-$\ell_2$]{
\includegraphics[width=0.48\textwidth]{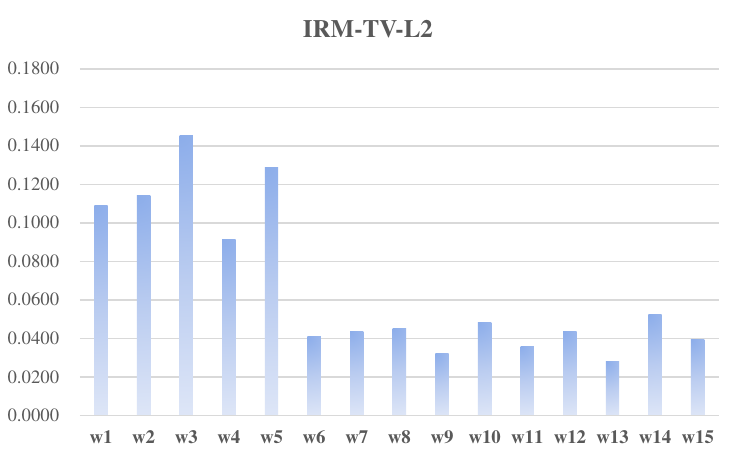}}
\hspace{4pt}\subfloat[IRM-TV-$\ell_1$]{
\includegraphics[width=0.48\linewidth]{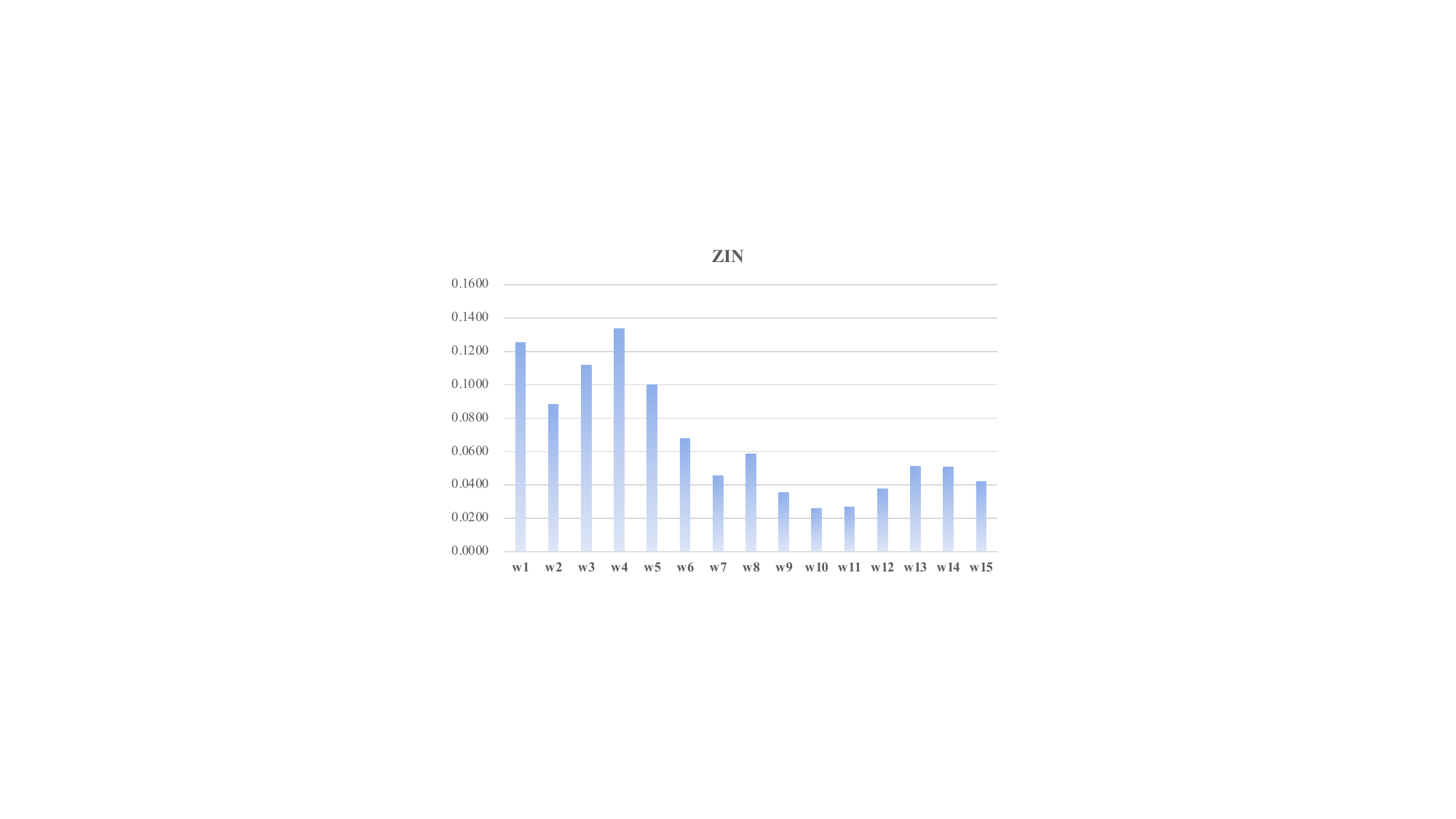}}\\
\subfloat[ZIN]{
\includegraphics[width=0.48\linewidth]{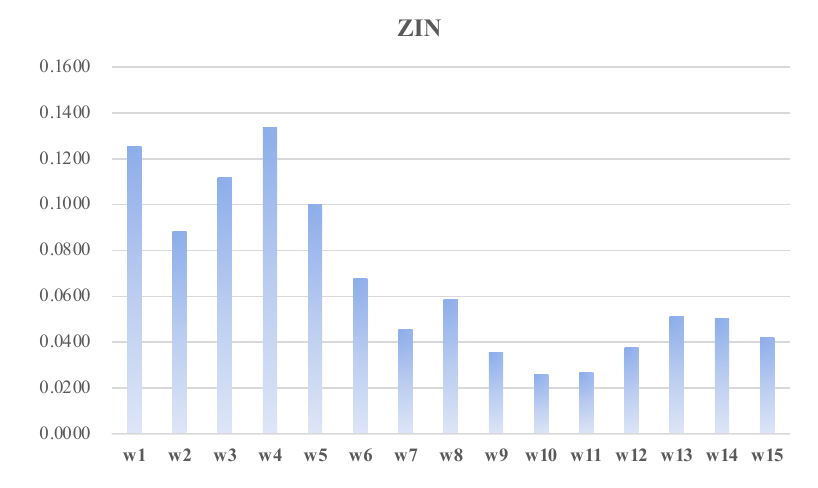}}
\subfloat[Minimax-TV-$\ell_1$]{
\hspace{4pt}\includegraphics[width=0.48\linewidth]{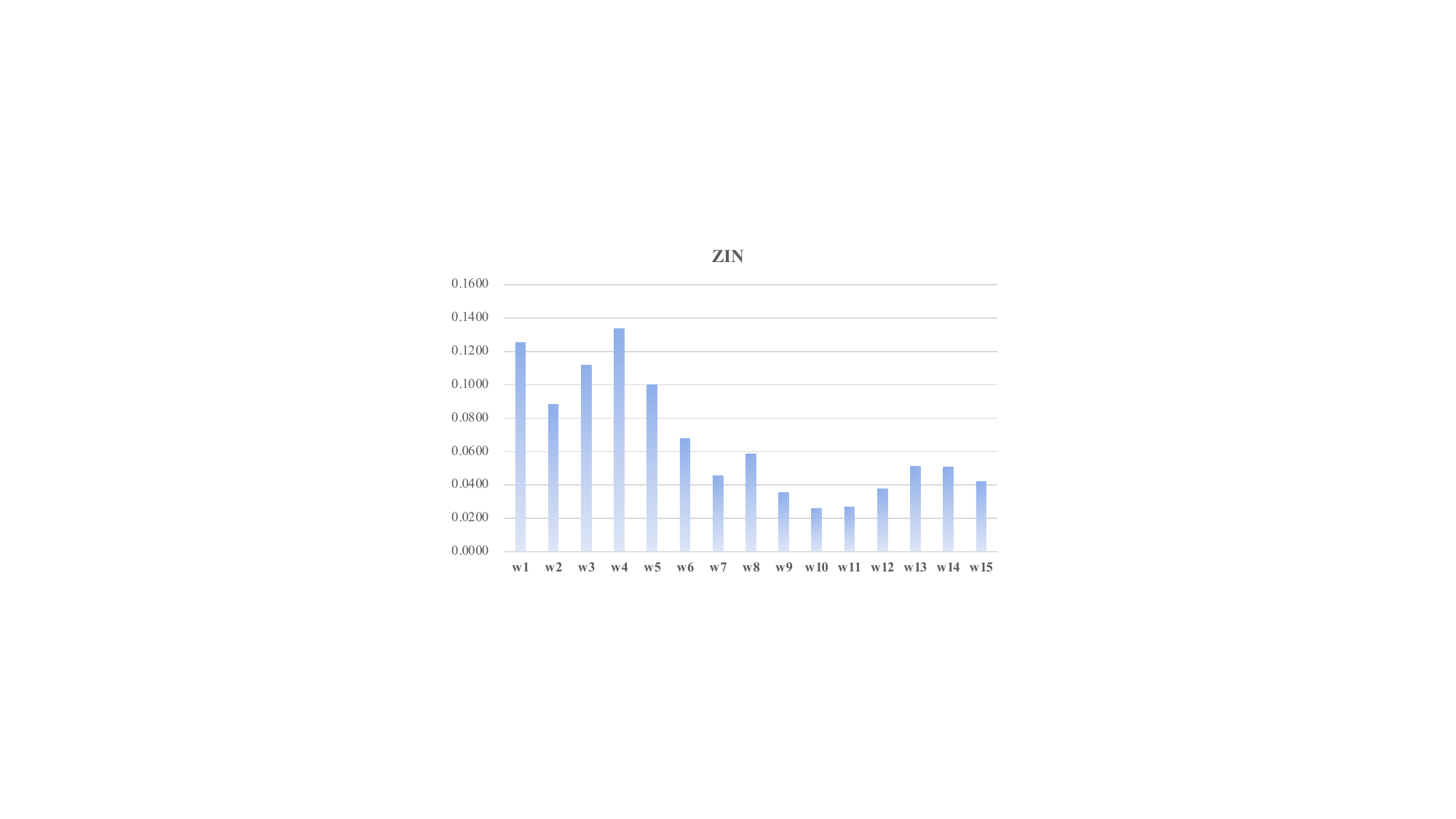}}
\caption{Normalized absolute values of feature weights for IRM-TV models in simulation study with $(p_{s}^{-}, p_{s}^{+}, p_{v}(t))=(0.999, 0.9, 0.8)$. \textbf{w1$\sim$w5} correspond to \textbf{invariant} features and \textbf{w6$\sim$w15} correspond to \textbf{spurious} features.}
\label{fig:invarfeat}
\end{figure*}

We take the simulation study with $(p_{s}^{-}, p_{s}^{+}, p_{v}(t))=(0.999, 0.9, 0.8)$ as an example. There are $15$ features in this experiment, where \textbf{No. 1$\sim$5} are \textbf{invariant} features and \textbf{No. 6$\sim$15} are \textbf{spurious} features. We run $10$ times of this experiment with different simulated samples and compute the normalized average absolute values of the feature weights, shown in Table \ref{tb:invarfeat} and Figure \ref{fig:invarfeat}. The invariant features of \ref{eqn:IRMv1wTVL1} and \ref{eqn:MinimaxTVL1} take up $64.88\%$ and $65.08\%$ of the total feature weights, respectively. In contrast, the invariant features of \ref{eqn:IRMv1wTVL2} and \ref{eqn:ZIN} take up only $58.88\%$ and $55.90\%$ of the total feature weights, respectively. Hence the TV-$\ell_1$ models extract more invariant features than the TV-$\ell_2$ models. Moreover, the gap between the invariant and spurious feature weights in Figures \ref{fig:invarfeat}(b) or \ref{fig:invarfeat}(d) is larger than that in Figures \ref{fig:invarfeat}(a) or \ref{fig:invarfeat}(c), respectively. For example,  the invariant feature weight w2$=0.0884$ and the spurious feature weights w6$=0.0679$ and w8$=0.0586$ for \ref{eqn:ZIN}, while w2$=0.1210$, w6$=0.0403$, and w8$=0.0446$ for \ref{eqn:MinimaxTVL1}, respectively. It indicates that \ref{eqn:MinimaxTVL1} enhances the invariant feature w2 while suppresses the spurious features w6 and w8.

\begin{table*}[h]
\caption{Normalized absolute values of feature weights for IRM-TV models in simulation study with $(p_{s}^{-}, p_{s}^{+}, p_{v}(t))=(0.999, 0.9, 0.8)$. \textbf{w1$\sim$w5} correspond to \textbf{invariant} features and \textbf{w6$\sim$w15} correspond to \textbf{spurious} features.}
\label{tb:invarfeat} 
\begin{center}
\begin{small}%\scriptsize
\begin{sc}
\scalebox{0.73}{\begin{tabular}{cccccc|cccccccccc}
\toprule
%\abovespace\belowspace
 Method & w1 &  w2 &  w3 &  w4 &  w5 &  w6 &  w7 &  w8 &  w9 &  w10 &  w11 &  w12 &  w13 &  w14 &  w15 \\
 \hline
 IRM-TV-$\ell_2$ &  0.1091 & 0.1143 & 0.1452 & 0.0912 & 0.1290 & 0.0412 & 0.0439 & 0.0454 & 0.0323 & 0.0483 & 0.0361 & 0.0438 & 0.0282 & 0.0526 & 0.0395 \\
 IRM-TV-$\ell_1$ &  0.1232 & 0.1292 & 0.1548 & 0.1090 & 0.1326 & 0.0329 & 0.0329 & 0.0446 & 0.0249 & 0.0371 & 0.0317 & 0.0386 & 0.0262 & 0.0482 & 0.0341 \\
 ZIN &  0.1253 & 0.0884 & 0.1118 & 0.1336 & 0.0999 & 0.0679 & 0.0454 & 0.0586 & 0.0353 & 0.0261 & 0.0267 & 0.0376 & 0.0510 & 0.0505 & 0.0419 \\
 Minimax-TV-$\ell_1$ &  0.1320 & 0.1210 & 0.1410 & 0.1353 & 0.1215 & 0.0403 & 0.0356 & 0.0446 & 0.0221 & 0.0279 & 0.0285 & 0.0341 & 0.0352 & 0.0395 & 0.0412 \\
\bottomrule
\end{tabular}}
\end{sc}
\end{small}
\end{center}
\end{table*}

\end{document}